\documentclass[journal]{IEEEtran}

\usepackage[table,xcdraw]{xcolor}
\usepackage{graphicx}
\usepackage{verbatim}
\usepackage{booktabs}
\usepackage{tabularx}
\usepackage[caption=false,font=footnotesize]{subfig}  
\usepackage{stfloats}

\usepackage{amsmath}
\usepackage{relsize}

\usepackage{makecell}
\usepackage{array}
\usepackage{makecell}
\usepackage{balance}

\newcommand{\ie}{{\em i.e., }}
\newcommand{\eg}{{\em e.g., }}

\newcommand{\myverb}{\fontsize{9}{48}\usefont{OT1}{lmtt}{b}{n}\noindent }

\usepackage{tcolorbox}

\newcommand{\EncoderAE}{\ensuremath{\mathrm{E}_{\text{AE}}}}
\newcommand{\EncoderVAE}{\ensuremath{\mathrm{E}_{\text{VAE}}}}
\newcommand{\EncoderAEEntity}{\ensuremath{\mathrm{E}_{\text{AE-entity}}}}
\newcommand{\EncoderAELimited}{\ensuremath{\mathrm{E}_{\text{AE-limited}}}}
\newcommand{\EncoderVAELimited}{\ensuremath{\mathrm{E}_{\text{VAE-limited}}}}
\newcommand{\Dataset}[2]{\ensuremath{\mathrm{#1}_{\text{#2}}}}

\ifCLASSINFOpdf
\else
\fi

\begin{document}
\title{Generalizable IoT Traffic Representations for Cross-Network Device Identification}

\author{Arunan~Sivanathan,
        David~Warren,
        Deepak~Mishra,
        Sushmita~Ruj,
        Natasha~Fernandes,
        Quan Z.~Sheng,
        Minh~Tran,\\
        Ben~Luo,
        Daniel~Coscia, 
        Gustavo~Batista   and~Hassan~Habibi~Gharakaheili
        \thanks{A.~Sivanathan, D.~Mishra, and H.~Habibi~Gharakaheili are with the School of Electrical Engineering and Telecommunications, University of New South Wales, Sydney, NSW 2052, Australia (e-mails: \{a.sivanathan, d.mishra, h.habibi\}@unsw.edu.au).}
        \thanks{S.~Ruj and G.~Batista are with the School of Computer Science and Engineering, University of New South Wales, Sydney, NSW 2052, Australia (e-mails: \{sushmita.ruj, g.batista\}@unsw.edu.au).}
        \thanks{D.~Warren,  N.~Fernandes and Q.Z.~Sheng are with the School of Computing, Macquarie University, Sydney, NSW 2109, Australia (e-mails: \{david.warren, natasha.fernandes, michael.sheng\}@mq.edu.au.}
        \thanks{M.~Tran, B.~Luo and D.~Coscia are with Information Sciences Division, Defence Science \& Technology Group, Edinburgh, SA 5111, Australia (e-mails: \{minh.tran7, ben.luo, daniel.coscia1\}@defence.gov.au).}                   
}


\maketitle

\begin{abstract}
Machine learning models have demonstrated strong performance in classifying network traffic and identifying Internet-of-Things (IoT) devices, enabling operators to discover and manage IoT assets at scale. However, many existing approaches rely on end-to-end supervised pipelines or task-specific fine-tuning, resulting in traffic representations that are tightly coupled to labeled datasets and deployment environments, which can limit generalizability. 
In this paper, we study the problem of learning generalizable traffic representations for IoT device identification. We design compact encoder architectures that learn per-flow embeddings from unlabeled IoT traffic and evaluate them using a frozen-encoder protocol with a simple supervised classifier. Our specific contributions are threefold.
(1) We develop unsupervised encoder--decoder models that learn compact traffic representations from unlabeled IoT network flows and assess their quality through reconstruction-based analysis. 
(2) We show that these learned representations can be used effectively for IoT device-type classification using simple, lightweight classifiers trained on frozen embeddings. 
(3) We provide a systematic benchmarking study against 
the 
state-of-the-art pretrained traffic encoders, showing that larger models do not necessarily yield more robust representations for IoT traffic.
Using more than 18 million real IoT traffic flows collected across multiple years and deployment environments, we learn traffic representations from unlabeled data and evaluate device-type classification on disjoint labeled subsets, achieving macro F1-scores exceeding 0.9 for device-type classification and demonstrating robustness under cross-environment deployment.

\end{abstract}


\IEEEpeerreviewmaketitle

\section{Introduction}

\IEEEPARstart{T}{he} rapid growth of Internet-of-Things (IoT) deployments has introduced a diverse and continuously evolving population of connected devices to enterprise, campus, and residential networks~\cite{IMC2020,22MudBrickJIoT,WoWMoM2024}. These devices range from cameras and voice assistants to sensors, power switches, and smart lighting systems, and often exhibit distinct communication patterns and operational lifecycles~\cite{BehavIoT2023,EuroSP25}. For network operators, maintaining accurate visibility into deployed IoT 
devices~\cite{CISA2025AssetInventory,NISTSP800-213}, including their types and behaviors, is essential for asset 
inventory \cite{cisaBOD23,omb2023fismaM2404}, policy enforcement \cite{ifipNetworkingMUDpolicy2024}, and security monitoring \cite{viakoo2024iotcrisis,2024sunblock}, particularly given the limited manageability and patching of many IoT devices \cite{paloalto2023iotbenchmark,bitdefender2024iot}. 

Passive network traffic analysis is widely used to support device identification in such environments, as it avoids active scanning, scales more naturally with network size, and does not require direct interaction with deployed devices. However, the heterogeneity of IoT devices and the continuous addition of new device types complicate robust identification across networks and deployment settings \cite{nozomi2024m2404}.
This challenge is particularly relevant for large enterprises, campuses, and service providers that manage IoT deployments at multiple sites, where frequent retraining of identification models for each environment is operationally expensive or impractical~\cite{ParticleIoTDvMgm2025,CiscoIoTDvMgm2025}.

A substantial body of prior work has studied IoT device identification using machine learning (ML) models applied to network traffic. Many approaches rely on end-to-end supervised pipelines \cite{18tmc,19sosrIoT,AuDI2019,20IoTJ1Class,20TNSMiotof,24tmlcn}, in which traffic representations are learned jointly with task-specific classifiers using labeled data collected from a fixed environment. While effective in controlled settings, such approaches are often evaluated under data-specific training assumptions and can experience performance degradation when applied to different device populations or network conditions \cite{Feamster:EuroSP24,NDSS2025}. 

More recent work has explored large pretrained traffic encoders \cite{ETBERT2022,YaTC023} to learn generic representations from large traffic corpora. Although these models can achieve strong in-dataset performance, they are computationally expensive, sensitive to design choices such as flow segmentation, and typically rely on task-specific fine-tuning (or re-training) to adapt to new settings. 
Recent studies have further shown that the apparent robustness of pretrained traffic models can be overstated under insufficiently rigorous evaluation protocols~\cite{SweetDangerSugar2025}, motivating a more careful examination of representation robustness.

In this work, we adopt a representation-centric perspective that decouples traffic representation learning from task-specific classification. We learn compact traffic representations from unlabeled IoT traffic using unsupervised encoder--decoder models, and subsequently use these representations as fixed inputs to simple supervised classifiers for identification of IoT device types. This separation enables representation learning to be performed once without labels, while downstream classifiers can be trained efficiently using fixed representations, without retraining or modifying the representation model. In this work, we focus on representation learning and evaluation methods, rather than on proposing new neural architectures.

Building on this perspective, we investigate two research questions. 
\textit{[RQ1]} Can traffic representations learned from unlabeled IoT traffic generalize across evolving IoT device populations and deployment environments (\S\ref{sec:traffic-representation}, \S\ref{sec:downstream-class}, and \S\ref{sec:benchmark})? 
\textit{[RQ2]} Does increasing model scale and generic pretraining inherently improve the robustness of traffic representations for IoT device identification, or can simpler representation-learning strategies achieve comparable generalizability (\S\ref{sec:benchmark})?


This paper makes the following 
main contributions.
\textbf{First}, we design and analyze unsupervised encoder--decoder architectures (\S\ref{sec:traffic-representation}) that learn compact traffic representations from unlabeled IoT network flows. Using reconstruction-based analysis, we study how architectural choices affect representation quality and stability. 
\textbf{Second}, we demonstrate that these learned representations can be effectively reused for IoT device-type classification (\S\ref{sec:downstream-class}) using simple supervised classifiers trained on frozen embeddings, showing that downstream performance is primarily governed by representation quality rather than classifier complexity.
\textbf{Third}, we provide a systematic benchmarking study (\S\ref{sec:benchmark}) that compares our encoder models against state-of-the-art pretrained traffic models under a unified frozen-encoder protocol and evaluates robustness across independent deployments. Our results show that larger pretrained models do not necessarily produce more robust representations for IoT device identification, and that latent-space regularization improves transferability.

Our study is grounded in more than 18 million real IoT traffic flows (\S\ref{sec:data}) collected over multiple years and deployment environments, reflecting realistic operational conditions.

\section{Related Work}\label{sec:prior}

We review prior work on IoT device identification and network traffic representation learning, and position our contributions relative to these efforts.

\vspace{1mm}
\noindent \textbf{Network Device Identification:} Research on identifying devices connected to a network broadly follows two approaches: active probing and passive traffic analysis.

Active methods have been widely used to discover traditional IT 
(Information Technology) assets (\ie personal computers and enterprise servers) by scanning hosts with TCP
(Transmission Control Protocol)/UDP 
(User Datagram Protocol) probes or issuing application-level requests and classifying devices based on their responses. Since uncontrolled active scans may disrupt resource-constrained IoT devices, prior work proposed lightweight~\cite {18iciafsIoT} or programmable~\cite{23noms} probing strategies tailored to IoT services. Work in~\cite{18iciafsIoT} demonstrated that IoT devices can be classified using TCP port scans, while the work in~\cite{23noms} introduced a programmable packet emitter that performs contextualized scans guided by early traffic profiles. Although these approaches reduce probing overhead, they still rely on active interaction and do not address continuous identification in heterogeneous deployments.   

Given operational risks of active probing, a large body of work has focused on passive IoT device identification using network traffic measurements (\ie packets and/or flows). Some studies rely on deterministic signatures extracted from observed traffic, such as application-layer parameters~\cite{23TIOT,WoWMoM2024}, cloud server identities~\cite{IoTFinder, Heidemann:IoTSec18,IMC2020}, or runtime Manufacturer Usage Description (MUD)-based fingerprints that combine transport-layer and endpoint information \cite{20TDSCmud}. These methods are interpretable and efficient, but their effectiveness can degrade when endpoints change, devices use shared backend infrastructure, or communication patterns evolve, often requiring manual updates or rule maintenance.

More recent work employs ML models trained on flow-level~\cite{18tmc,AuDI2019,20IoTJ1Class,20TNSMiotof,NDSS2025} or packet-level features~\cite{23globecom,24tmlcn} to learn statistical fingerprints of device behavior. Such approaches can capture subtle behavioral patterns, even with encrypted payloads, but typically require labeled training data and careful validation. 

Recent work \cite{NDSS2025} systematically evaluated the robustness of ML-based IoT device identification under realistic operational conditions. The authors showed that flow-level classifiers can experience substantial performance degradation across modes of operation, time, and deployment environments. While this study provides important evidence of robustness limitations, it focuses on classifier behavior and does not examine representation learning or the reuse of frozen traffic embeddings across environments, which are central to our work. 

\vspace{1mm}
\noindent \textbf{Representation Learning for Network Traffic:} A growing body of research explores representation learning for network traffic using self-supervised or pretraining-based paradigms. ET-BERT~\cite{ETBERT2022} adapts transformer-based language models to encrypted traffic by treating packet bytes as token sequences and pretraining encoders that are subsequently fine-tuned for downstream classification tasks. YaTC \cite{YaTC023} extends this line of work by proposing a masked autoencoder-based Traffic Transformer that explicitly models byte-, packet-, and flow-level structures, reporting strong in-dataset and few-shot performance after supervised fine-tuning. NetMamba~\cite{Netmamba2024} replaces attention mechanisms with a state-space model to improve efficiency while retaining a pretraining-and-fine-tuning pipeline.

Although these approaches improve traffic classification accuracy, they are typically evaluated after dataset-specific fine-tuning, leaving open the question of whether representations learned from unlabeled traffic can be reused in a frozen form across deployments. Recent studies have questioned the robustness of large pretrained traffic encoders, showing that the reported gains can be inflated by dataset artifacts and evaluation choices~\cite{SweetDangerSugar2025}, and that weaknesses in learned representations become apparent when examined beyond downstream accuracy~\cite{NetFoundationModels2025}. These findings highlight the need for careful assessment of representation quality under realistic deployment settings.

Representation learning has also been applied to traffic generation. NetFlowGen~\cite{NetFlowGen2024} proposed a generative framework that learns flow-level representations to synthesize realistic NetFlow records, primarily for data augmentation and DDoS 
(Distributed Denial-of-Service) detection. While this work 
demonstrated that learned latent variables can capture statistical properties of network flows, it evaluated representations through generation quality rather than reuse for inference. In contrast, our work focuses on encoder-side representations and explicitly evaluates their reusability and robustness across datasets and deployment environments for IoT device identification.

\vspace{1mm}
\noindent \textbf{Our Novelty:} This work involves two key innovations in the context of ML-based IoT traffic analysis. \textit{First}, we explicitly formulate IoT device identification as a representation generalizability problem. While prior ML-based approaches evaluate performance under dataset-specific training or fine-tuning, our study focuses on whether traffic representations learned from unlabeled data can remain reusable as device populations and deployment environments evolve.
\textit{Second}, our results provide empirical evidence that robustness in IoT traffic representations depends more on representation design and latent regularization than on model scale or generic pretraining alone. Through systematic cross-dataset and cross-environment evaluation, we show that representations learned with latent regularization remain substantially more stable under deployment shift than deterministic encoders without explicit probabilistic latent modeling.


\begin{figure*}[t!]
    \centering
   \includegraphics[width=.90\linewidth]{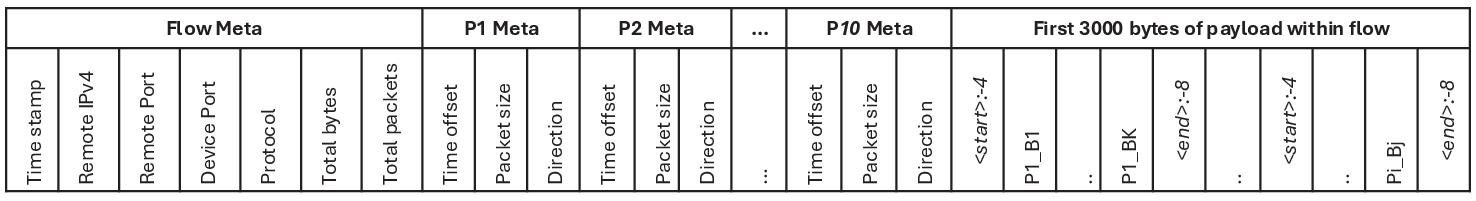}
    \vspace{-1mm}
    \caption{Structure of a Custom Flow.}
    \label{fig:CustomFlow}
    \vspace{-5mm}
\end{figure*}

\section{IoT Traffic Datasets}
\label{sec:data}
This study evaluates the generalizability of learned traffic representations using three real IoT traffic datasets collected from distinct environments and across different time periods. All datasets are processed into a unified \textit{Custom Flow} representation to ensure consistency across experiments.

\vspace{1mm}
\noindent \textbf{Raw traffic representation:}
In our previous work \cite{25TNSE}, we introduced a raw network traffic structure termed  \textit{Custom Flow}, which transforms packet captures (PCAPs) into enriched flow records that retain structured traffic information relevant to device behavior. A Custom Flow aggregates a compact set of flow-level metadata, packet-level timing and direction information, and selected payload bytes from the initial packets of a communication.
A Custom Flow record is uniquely identified by a five-tuple and is capped at a maximum duration of one-minute. 
Long-lived connections (common in some IoT cloud communications) are segmented into consecutive one-minute flow records that share the same identifier. This timeout represents an operational trade-off that enables near-real-time inference while preserving session-level semantics through consecutive flow segmentation.

Each Custom Flow record is represented as a fixed-length vector of dimension $1 \times 3040$, serving as input to all encoder models in this paper, as shown in Fig.~\ref{fig:CustomFlow}. 
Specifically, each flow includes: relative flow time; remote IPv4 address segmented into four octets; transport-layer protocol; remote and device-side transport-layer port numbers; byte count and packet count; and three metadata arrays capturing the first ten packets’ time offsets, packet sizes, and packet directions.
To capture early behavioral fingerprints, up to 3,000 bytes of payload from these packets are extracted and concatenated sequentially, with special marker values ($-4$ and $-8$) used as delimiters and padding represented by $-255$ to avoid ambiguity with valid byte values.
This fixed-dimensional design enables direct ingestion by neural network models without additional preprocessing. Following prior observations that device-initiated traffic carries the most stable and discriminative behavioral signals for IoT device identification~\cite{25TNSE}, we restrict our analysis to \textit{unidirectional outgoing flows} (device-to-network) to reduce ambiguity introduced by externally driven responses.

\vspace{1mm}
\noindent \textbf{\textrm{DATA16}:} This dataset was collected in 2016 from a smart-home testbed populated with first-generation consumer IoT devices. It comprises 60 days of continuous traffic traces recorded between 23 September and 22 November 2016, yielding approximately 3.1 million unidirectional flows from 19 device types. The dataset is publicly available~\cite{DryadCustomFlows}. Following the temporal ordering of the traces, we partition the data into a 7-day training (\Dataset{DATA16}{train}), a 5-day validation set \Dataset{DATA16}{val}), and the remaining 48 days forming the test set (\Dataset{DATA16}{test}).

\vspace{1mm}
\noindent \textbf{\textrm{DATA25v1}:} This dataset was collected in 2025 from an updated testbed comprising IoT devices representative of the 2024--2025 consumer market. It contains 57 days of packet traces from 11 February to 11 April 2025, yielding approximately 7.7 million flows across 18 device types. Among these devices, two correspond to newer hardware generations of device families present in \Dataset{DATA16}{} (\eg Amazon Echo $\rightarrow$ Echo Show~5, TP‑Link plug $\rightarrow$ Tapo plug), while fifteen represent entirely new consumer IoT device types. In addition, the dataset includes a non‑IoT ``\textit{IT}'' class, comprising general-purpose computing devices (\eg laptops, smartphones, and tablets) that differ from those in \Dataset{DATA16}{}. We chronologically split the dataset into an 8-day training set (\Dataset{DATA25v1}{train}), a 3-day validation set (\Dataset{DATA25v1}{val}), and the remaining 46 days as the test set (\Dataset{DATA25v1}{test}). 

\vspace{1mm}
\noindent \textbf{\textrm{DATA25v2}:} This dataset was collected during the same period and using the same collection methodology as \Dataset{DATA25v1}{}, but from a separate university laboratory environment. 
Although device categories overlap with $\Dataset{DATA25v1}{}$, the physical setup, network topology, and background traffic differ, making this dataset well-suited for evaluating cross-environment generalization. It comprises approximately 8.8 million flows from 10 device types. In our experiments, all samples from this dataset are used exclusively as a test set (\Dataset{DATA25v2}{test}).

We will publicly release \Dataset{DATA25v1}{} and \Dataset{DATA25v2}{} to support reproducibility and future research, to be available at \cite{github26}. 
While many public traffic datasets exist, we focus on IoT-specific deployments collected across different environments and time periods, as our goal is to study representation generalizability under realistic IoT operational conditions rather than across unrelated traffic domains.

Across all experiments, the datasets are partitioned to strictly separate representation learning from downstream evaluation. Encoder models are trained in an unsupervised manner using unlabeled traffic only, while supervised classifiers are trained and evaluated on disjoint labeled subsets. No flow instance is shared between representation learning and downstream evaluation to prevent information leakage.

\section{Network Traffic Representation}
\label{sec:traffic-representation}

This section presents our approach for transforming high-dimensional raw network traffic into compact semantic embeddings suitable for downstream IoT device classification tasks. We begin by motivating the need for learned traffic representations (\S\ref{sec:why}), followed by the design of our baseline autoencoder (\S\ref{sec:ae-arch}), hyperparameter optimization and reconstruction analysis (\S\ref{sec:ae-hparam}), an extension incorporating entity embeddings for categorical features (\S\ref{sec:entity-embeddings}), and finally a variational autoencoder (VAE) formulation that improves generalization to unseen network patterns (\S\ref{sec:vae-arch}).

\begin{figure*}[t!]
    \centering
   \includegraphics[width=.90\linewidth]{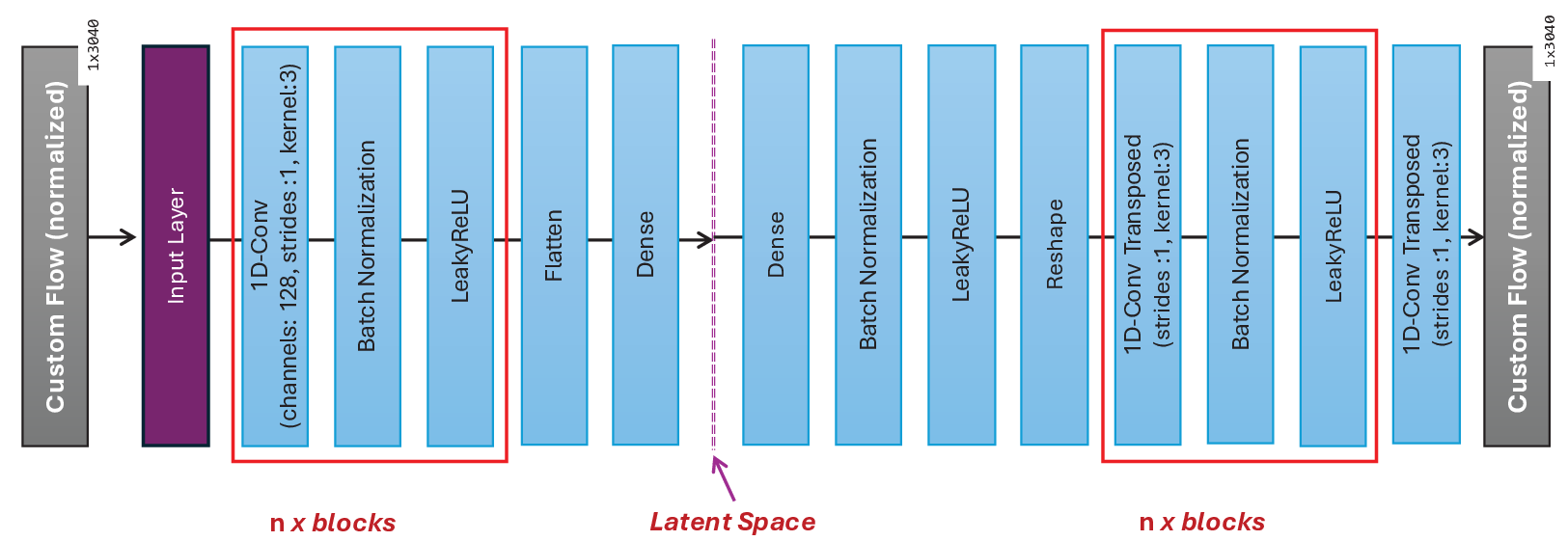}
    \vspace{-1mm}
    \caption{Architecture of our auto-encoder.}
    \label{fig:networkdiagram}
    \vspace{-3mm}
\end{figure*}

\subsection{Motivation: Why Traffic Embeddings?}\label{sec:why}

Prior work, including ours~\cite{25TNSE}, commonly feeds raw network traffic directly into classification models. However, such traces may contain noise such as timestamp jitter, packet-level variability, and transport-protocol behaviors (\eg handshakes, retransmissions, and record framing), all of which can obscure meaningful behavioral structure. Moreover, raw traffic representations are inherently high-dimensional, resulting in significant computational overhead during training and/or fine-tuning.

To address these limitations, we convert raw flows into \textit{low-dimensional embeddings} that preserve key semantic characteristics of network behavior. These embeddings reduce computational cost,  improve generalization by encouraging models to learn abstract traffic patterns, and facilitate reuse across environments without retraining from scratch.

Although several embedding-based approaches exist, such as transformers, autoencoders (AEs), and variational autoencoders (VAEs), we begin with an AE-based framework due to its architectural simplicity and effectiveness in learning compact representations. We then systematically extend this baseline to examine design choices that improve robustness.

\begin{figure*}[!bt]
    \centering
    \subfloat[$\mathbf{n}=1$.]{
        \includegraphics[width=0.45\textwidth]{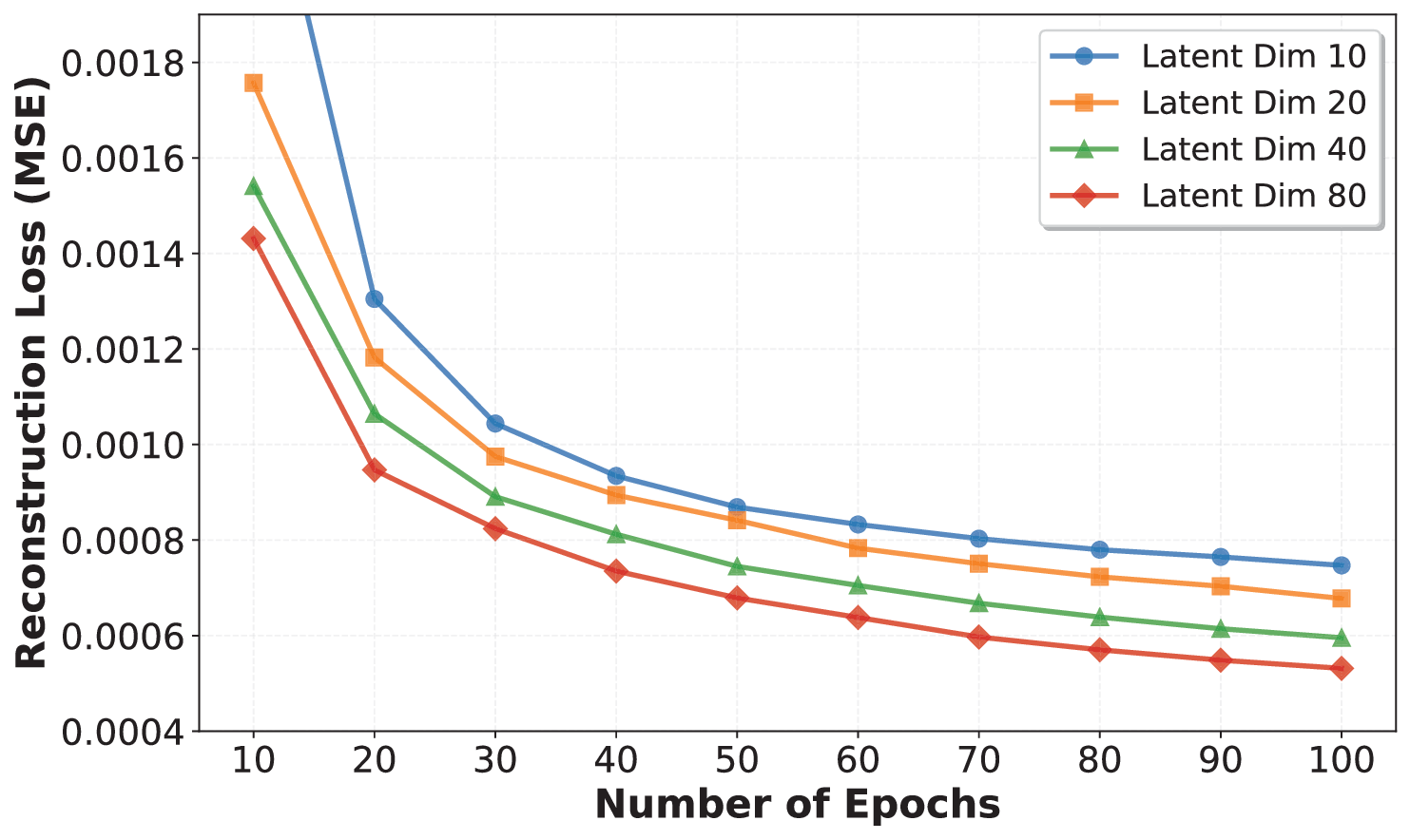}
        \label{fig:reconstruction-loss-block1}
    }
    \hfill
    \subfloat[$\mathbf{n}=2$.]{
        \includegraphics[width=0.45\textwidth]{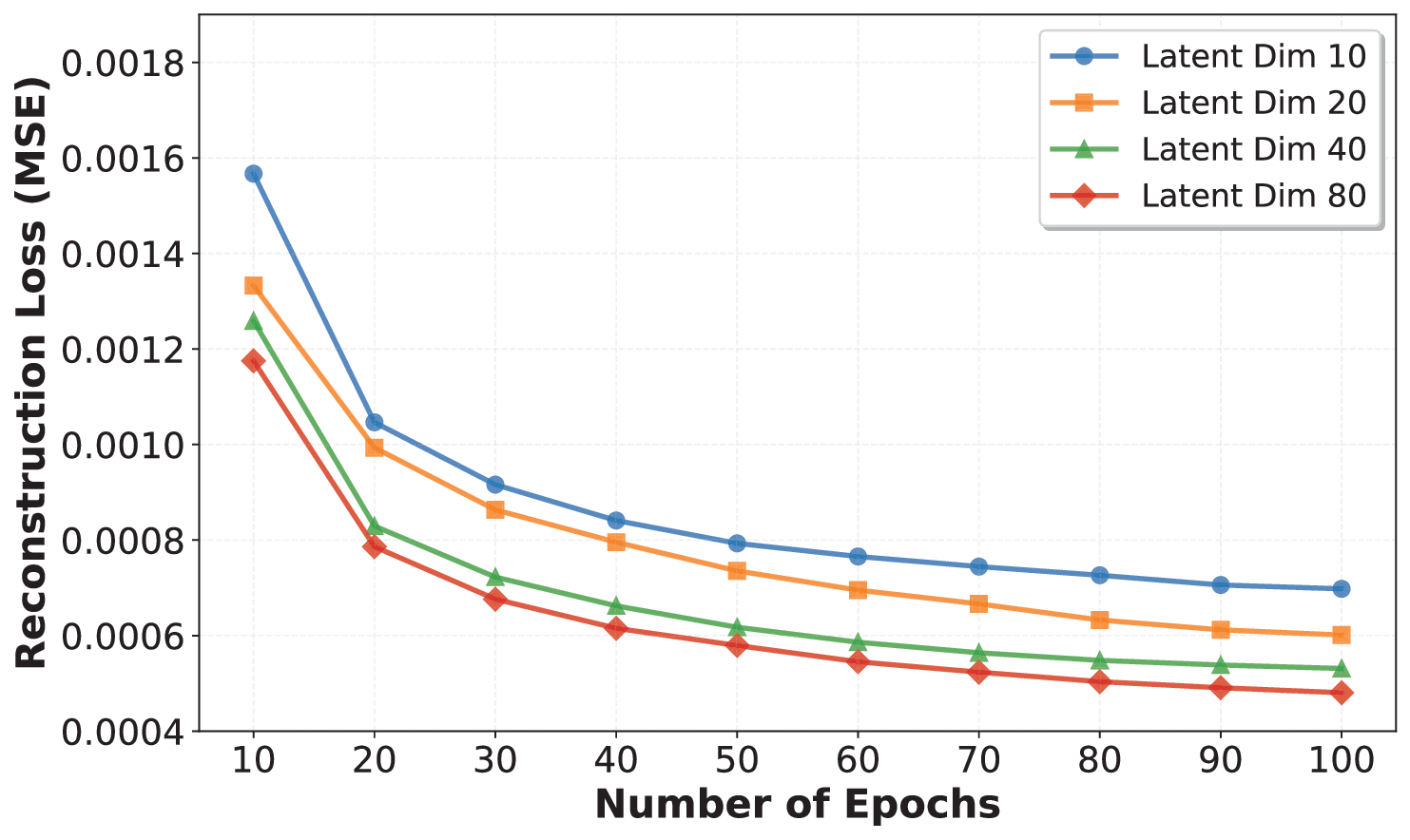}
        \label{fig:reconstruction-loss-block2}
    }
    \vspace{1mm}
    \subfloat[$\mathbf{n}=4$.]{
        \includegraphics[width=0.45\textwidth]{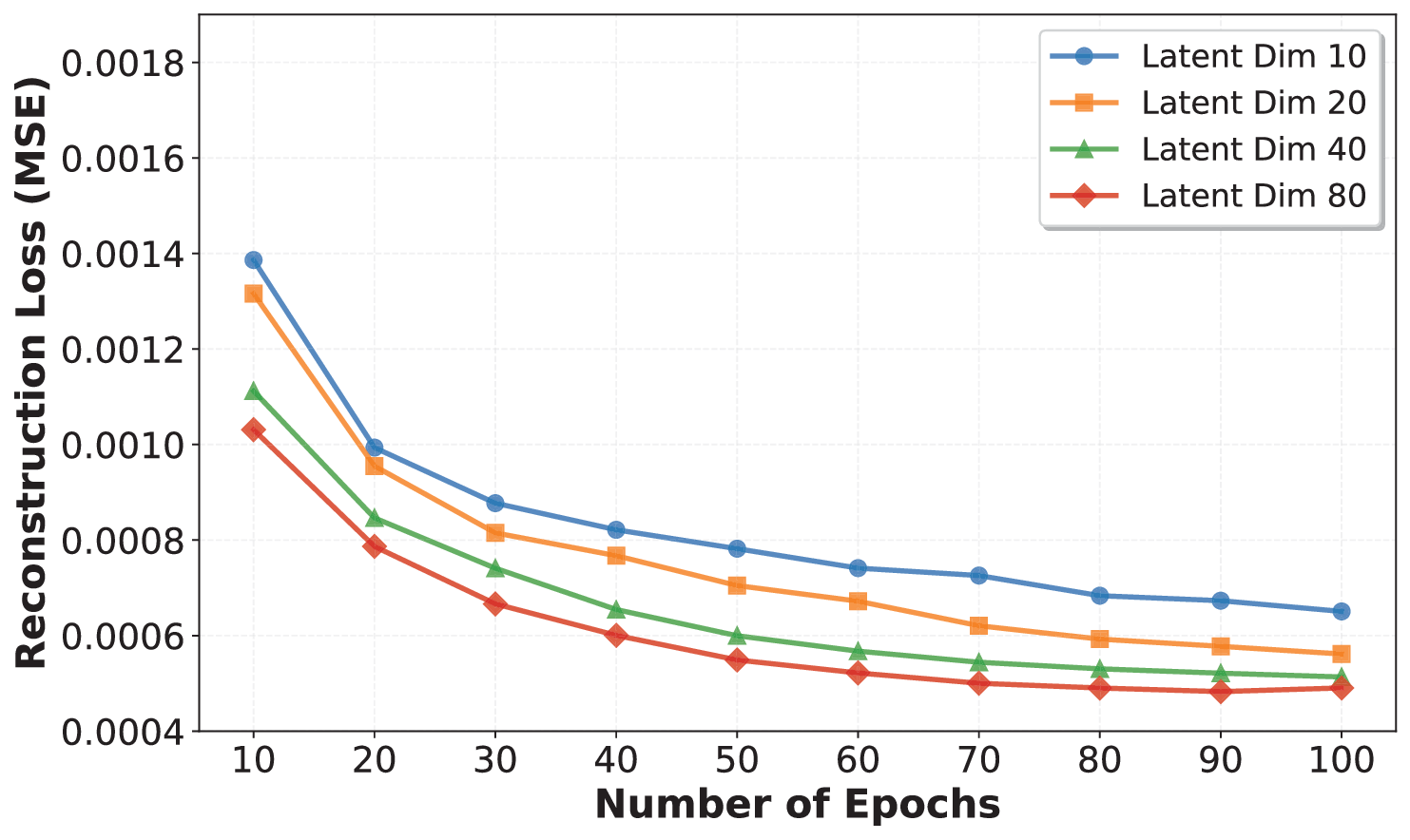}
        \label{fig:reconstruction-loss-block4}
    }
    \hfill
    \subfloat[$\mathbf{n}=8$.]{
        \includegraphics[width=0.45\textwidth]{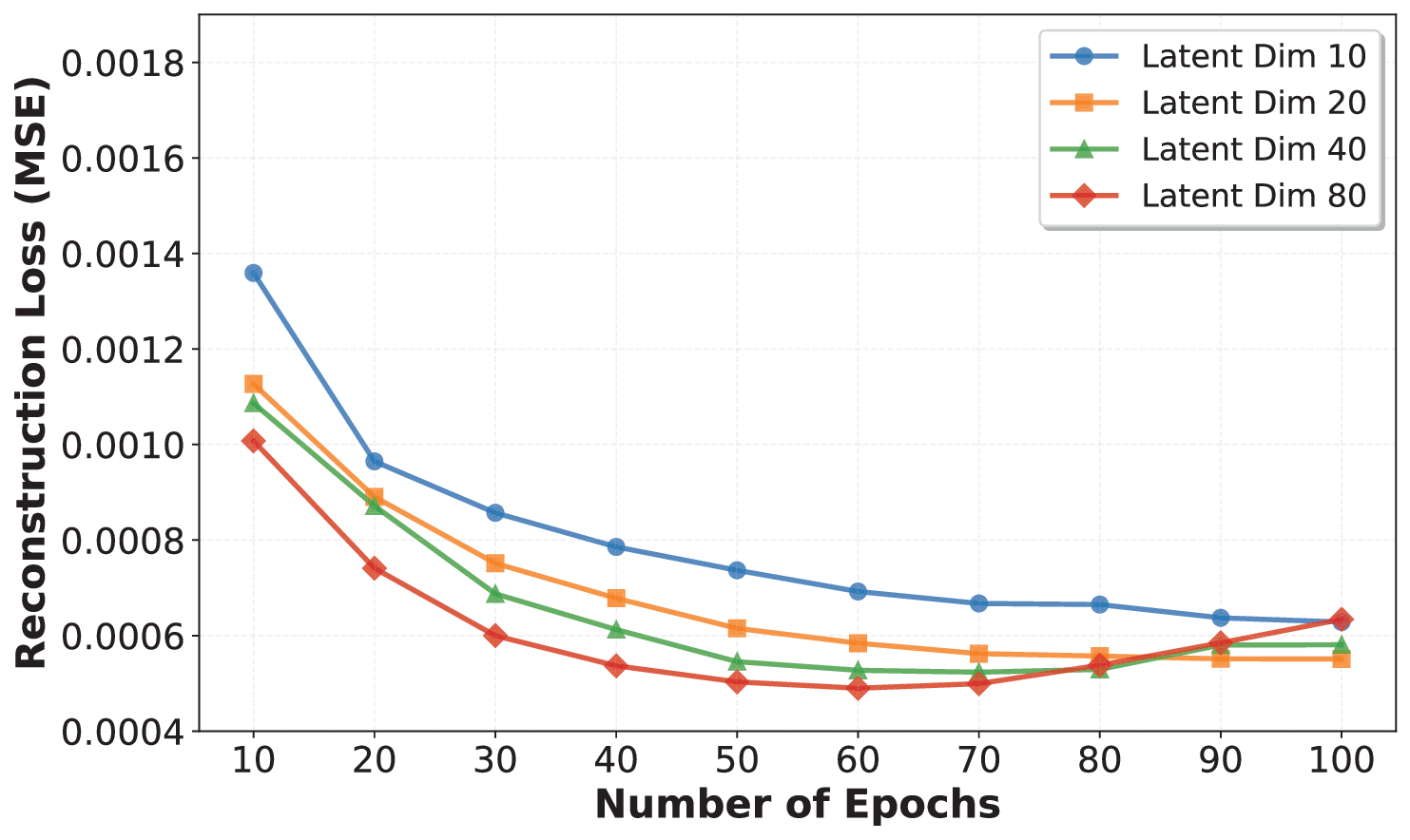}
        \label{fig:reconstruction-loss-block8}
    }

    \caption{Mean squared error (MSE) of reconstruction on the validation set as a function of training epochs for different autoencoder configurations. Each subplot corresponds to a distinct encoder depth $\mathbf{n}$, while colored curves within each subplot represent different latent dimensions $\mathbf{i} \in \{10, 20, 40, 80\}$.}
    \label{fig:reconstruction-loss}
    \vspace{-3mm}
\end{figure*}

\subsection{Baseline Autoencoder Architecture}
\label{sec:ae-arch}

To extract semantically meaningful representations from high-dimensional network flows, we design a convolutional autoencoder that compresses Custom Flow vectors of size $1 \times 3,040$ into a compact latent space. 
Prior to training, every field of the Custom Flow, namely relative flow time; the four remote IPv4 octets; transport protocol; remote and device port numbers; packet and byte counts; time offsets, sizes and directions for the first 10 packets; and up to 3,000 payload bytes, is normalized to $[0,1]$ to facilitate stable and efficient gradient-based optimization (special marker values such as padding bytes and packet delimiters are mapped to $[0,0.5)$ while valid data are mapped to $[0.5,1]$).
The autoencoder is trained unsupervised, and no device labels are provided during representation learning.
The autoencoder follows a standard encoder–decoder structure, where an \textit{encoder} maps the normalized input flow to a latent representation and a \textit{decoder} reconstructs the normalized flow from this representation.

The architecture is inspired by standard convolutional autoencoder designs for unsupervised representation learning over structured one-dimensional inputs~\cite{ICANN2011}. The Custom Flow representation preserves the ordering of packet-level metadata and payload segments while enabling one-dimensional convolutions to capture local dependencies in the traffic sequence. Multiple convolutional blocks are stacked to extract increasingly abstract features while keeping the model compact.

As shown in Fig.~\ref{fig:networkdiagram}, the encoder 
comprises $\mathbf{n}$ convolutional blocks (indicating depth), each containing a one-dimensional convolutional (1D-Conv) layer with 128 channels, kernel size 3, and unit stride, followed by batch normalization and a LeakyReLU activation. The resulting feature maps are flattened and passed through a dense layer to obtain a latent representation of dimension $\mathbf{i}$.

The decoder mirrors the encoder. A dense layer expands the latent vector, followed by batch normalization, LeakyReLU activation, and a reshape operation. This is followed by $\mathbf{n}$ transposed convolutional blocks, concluding with a 1D transposed convolution that reconstructs the original $1 \times 3,040$ flow vector.
Batch normalization and LeakyReLU activations improve training stability and robustness when learning from heterogeneous and sparse traffic features. 

\begin{figure}[t!]
    \centering
    \includegraphics[width=1\linewidth]{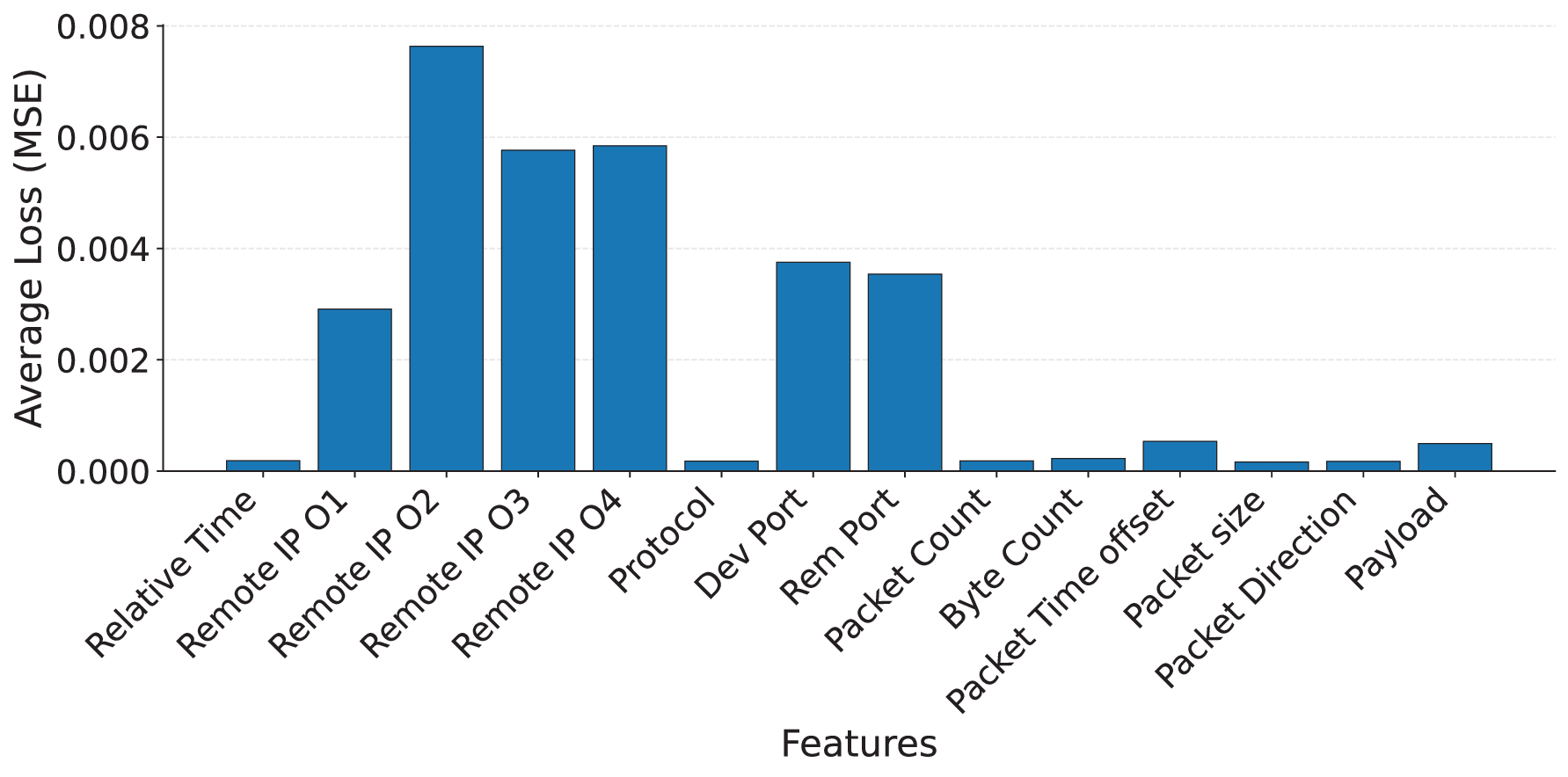}
    \caption{MSE of reconstructed numerical features on the validation set.}
    \label{fig:feature-loss}
    \vspace{-5mm}
\end{figure}

Autoencoder models in this section are trained exclusively using \Dataset{DATA16}{train}, with \Dataset{DATA16}{val} used only for configuration selection and reconstruction analysis. 
To maintain a balanced and diverse traffic profile across device types, we cap the number of training instances at 5,000 flows per device. To prevent data leakage, \Dataset{DATA16}{test} is reserved solely for downstream classification in \S\ref{sec:downstream-class}.

In what follows, we analyze a deterministic autoencoder to establish baseline reconstruction behavior before introducing a variational extension to study the impact on latent-space regularization.

\subsection{Hyperparameter Search and Reconstruction Analysis}
\label{sec:ae-hparam}

We determine optimal autoencoder hyperparameters by training on  \Dataset{DATA16}{train} and validating on \Dataset{DATA16}{val}. No samples from \Dataset{DATA16}{test} are used during encoder training or model selection. 

We perform a grid search over encoder depth $\mathbf{n} \in \{1, 2, 4, 8\}$ and latent space dimension $\mathbf{i} \in \{10, 20, 40, 80\}$. Reconstruction loss is measured as the mean squared error (MSE) across all 3,040 features (of the Custom Flow) and evaluated on the validation set. 

Fig.~\ref{fig:reconstruction-loss} shows that 
larger latent dimensions ($\mathbf{i}=40$ and $\mathbf{i}=80$) consistently achieve lower reconstruction loss across all encoder depths. While $\mathbf{n}=4$ and $\mathbf{n}=8$ perform 
the 
best overall, the $\mathbf{n}=8$ configuration becomes unstable after approximately 60 epochs, indicating overfitting. 
We 
therefore adopt $\mathbf{n}=4$ and $\mathbf{i}=40$ for the remainder of this study, as increasing the latent dimension to $\mathbf{i}=80$ yields only marginal additional reduction in reconstruction loss while doubling the embedding size and downstream model complexity.

\begin{figure}[t!]
    \centering
    \subfloat[A representative TCP flow from Amazon Echo (input vs. reconstruction).]{
        \includegraphics[width=0.45\linewidth]{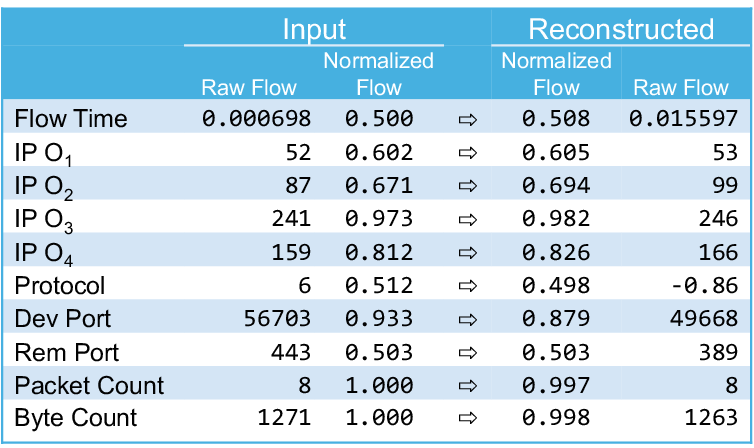}
        \label{fig:reconstruction-tcp}
    }
    \hspace{1mm}
    \subfloat[A representative UDP flow from LiFX lightbulb (input vs. reconstruction).]{
        \includegraphics[width=0.45\linewidth]{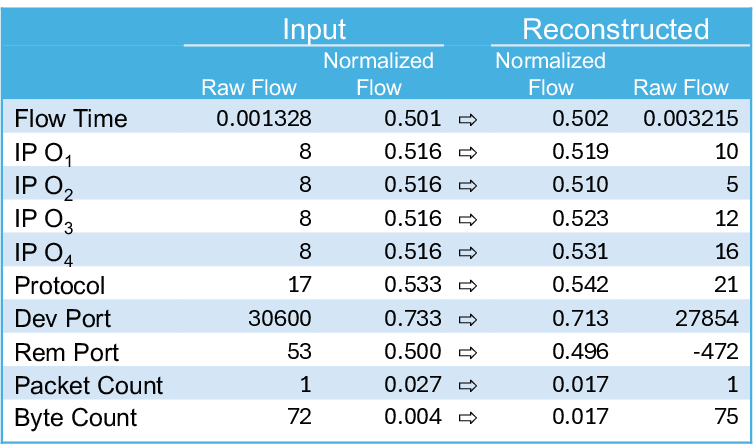}
        \label{fig:reconstruction-udp}
    }
    \caption{Reconstruction of two randomly selected flows from the validation set \Dataset{DATA16}{val}: (a) a sample TCP flow and (b) a sample UDP flow, showing raw and normalized feature values alongside their reconstructed counterparts.}
    \label{fig:reconstruction-tcp-udp}
    \vspace{-5mm}
\end{figure}

To examine reconstruction behavior across feature groups, Fig.~\ref{fig:feature-loss} reports the average reconstruction loss per feature on the validation set. 
Remote IP octets, particularly O2--O4, exhibit higher loss values 
than port features in the normalized domain. 
Fig.~\ref{fig:reconstruction-tcp-udp} provides a concrete example of how the autoencoder reconstructs individual flows, showing that key structural and semantic properties of the traffic are preserved even when exact numeric values differ.
Manual inspection of representative TCP and UDP flows (Fig.~\ref{fig:reconstruction-tcp-udp})  reveals that this effect arises from normalization: 
IP octets mapped from $[0,255]$ to $[0.5,1]$ amplify small raw deviations, whereas port numbers mapped from $[1,65535]$ exhibit larger absolute reconstruction errors in the raw domain.


\begin{figure}[t!]
    \centering
    \includegraphics[width=1\linewidth]{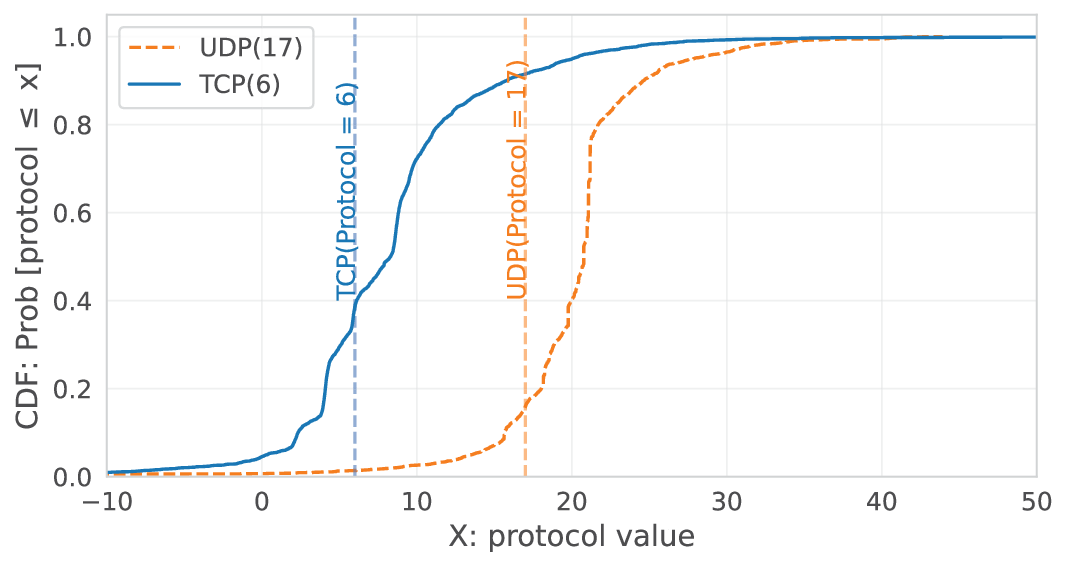}
    \caption{Reconstructed values of the protocol feature in \Dataset{DATA16}{val} from our baseline autoencoder.}
    \label{fig:numeric-proto-reconstruction}
\end{figure}

Further analysis of reconstructed protocol values (see Fig.~\ref{fig:numeric-proto-reconstruction}) shows that 
although reconstructed protocol values do not exactly match original identifiers (TCP = 6, UDP = 17), TCP and UDP flows remain largely separable. Approximately $80$\% of reconstructed TCP flows fall below a value of $11.5$, while fewer than $3$\% of UDP flows do so; conversely, more than $80$\% of reconstructed UDP flows exceed a value of $21$, compared to fewer than $4$\% of TCP flows.

These results suggest a potential limitation of treating all features as purely numerical. Protocol and port numbers are inherently categorical, as their numeric ordering does not convey semantic meaning  (\eg port 80 is not semantically closer to port 79 than to port 443). Motivated by these observations, we explore an alternative in which Protocol, Device Port, and  Remote Port are treated as categorical features in the following subsection. We do not apply the same treatment to the IP address octets, which retain meaningful numerical structure. 
For example, addresses in the {\myverb{8.8.8.x}} range are closer to {\myverb{8.8.9.x}} than to {\myverb{10.0.0.x}}. 

We emphasize that reconstruction analysis serves as a diagnostic tool to understand representation behavior; the effectiveness of the learned embeddings is evaluated explicitly in downstream classification.

\subsection{Categorical Feature Representation by Entity Embeddings}\label{sec:entity-embeddings}

Transport-layer protocol and port usage constitute core behavioral primitives of IoT devices and have been shown to be highly discriminative for device identification in prior work~\cite{25TNSE}. In this subsection, we introduce an entity-embedding variant of our autoencoder to examine whether the learned latent representations preserve these discrete behavioral semantics.

Entity embeddings provide a compact alternative to one-hot encoding for high-cardinality categorical features. By mapping categories to learned vectors, the model can capture behavioral similarity between categories while reducing dimensionality.

\begin{figure}[t!]
    \centering
    \includegraphics[width=0.975\linewidth]{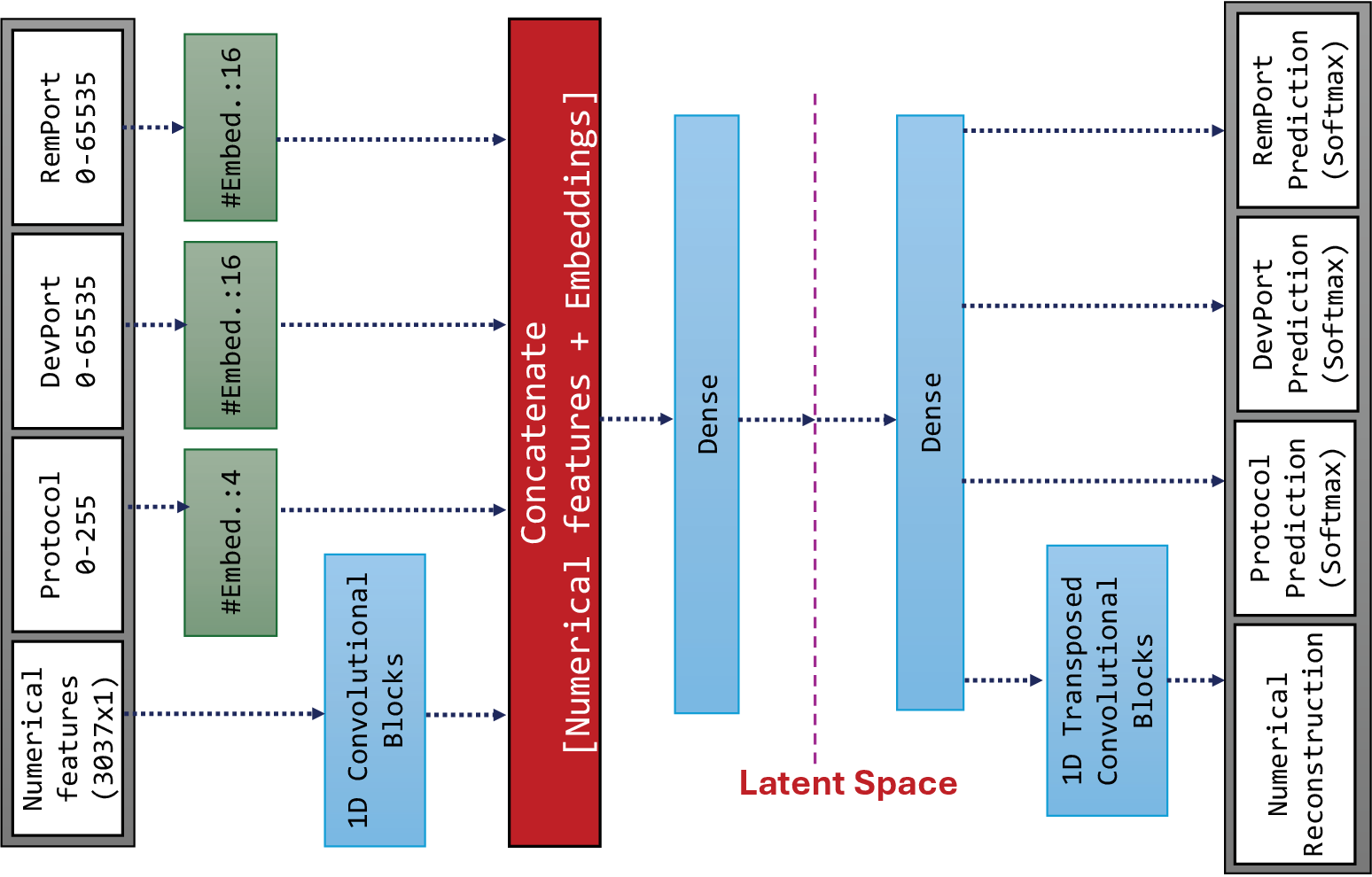}
    \caption{Hybrid encoder–decoder architecture with entity embeddings for categorical features and convolutional processing of numerical Custom Flow attributes.}
    \label{fig:architecture-entity-embeddings}
    \vspace{-5mm}
\end{figure}

As shown in Fig.~\ref{fig:architecture-entity-embeddings}, three features, namely Protocol, Device Port, and Remote Port, are processed through dedicated embedding layers, while the remaining 3,037 numerical features are handled by convolutional blocks shown in Fig.~\ref{fig:networkdiagram}. 
Embedding dimensions are selected using heuristic $d = \sqrt[4]{C}$ \cite{EntityEmbed2016}, where $C$ denotes the feature cardinality, resulting in dimensions of 16, 16, and 4 used for Device Port, Remote Port, and Protocol, respectively. 

In the decoder, numerical features are reconstructed via transposed convolutions, while categorical features are reconstructed via softmax outputs optimized with sparse categorical cross-entropy. The total loss combines numerical and categorical components with appropriate scaling: 
\begin{equation}
    L_{\text{total}} = L_{\text{cnn}} + 0.001 \times (L_{\text{Proto}} + L_{\text{RemPort}} + L_{\text{DevPort}}),
\end{equation}
where the scaling factor 0.001 balances the contribution of the three categorical losses against the approximately 3000 numerical features.

Using entity embeddings, the exact reconstruction rates for Protocol, Remote Port, and Device Port reach $99.4$\%, $83.3$\%, and $40.9$\%, respectively. Lower reconstruction for Device Port is expected due to ephemeral port usage on the device side. The correctly reconstructed $40.9$\% correspond to 28 frequently occurring ports associated with device services (\eg {\myverb{UDP/1900}} for SSDP, {\myverb{UDP/56700}} for LIFX local control) and standard client-side ports (\eg {\myverb{UDP/68}} for DHCP, {\myverb{UDP/5353}} for mDNS). 
Importantly, reconstruction accuracy is not a performance objective in itself, but a diagnostic to assess whether the latent space preserves meaningful device behavior.

\begin{figure}[t!]
    \centering
    \includegraphics[width=0.96\linewidth]{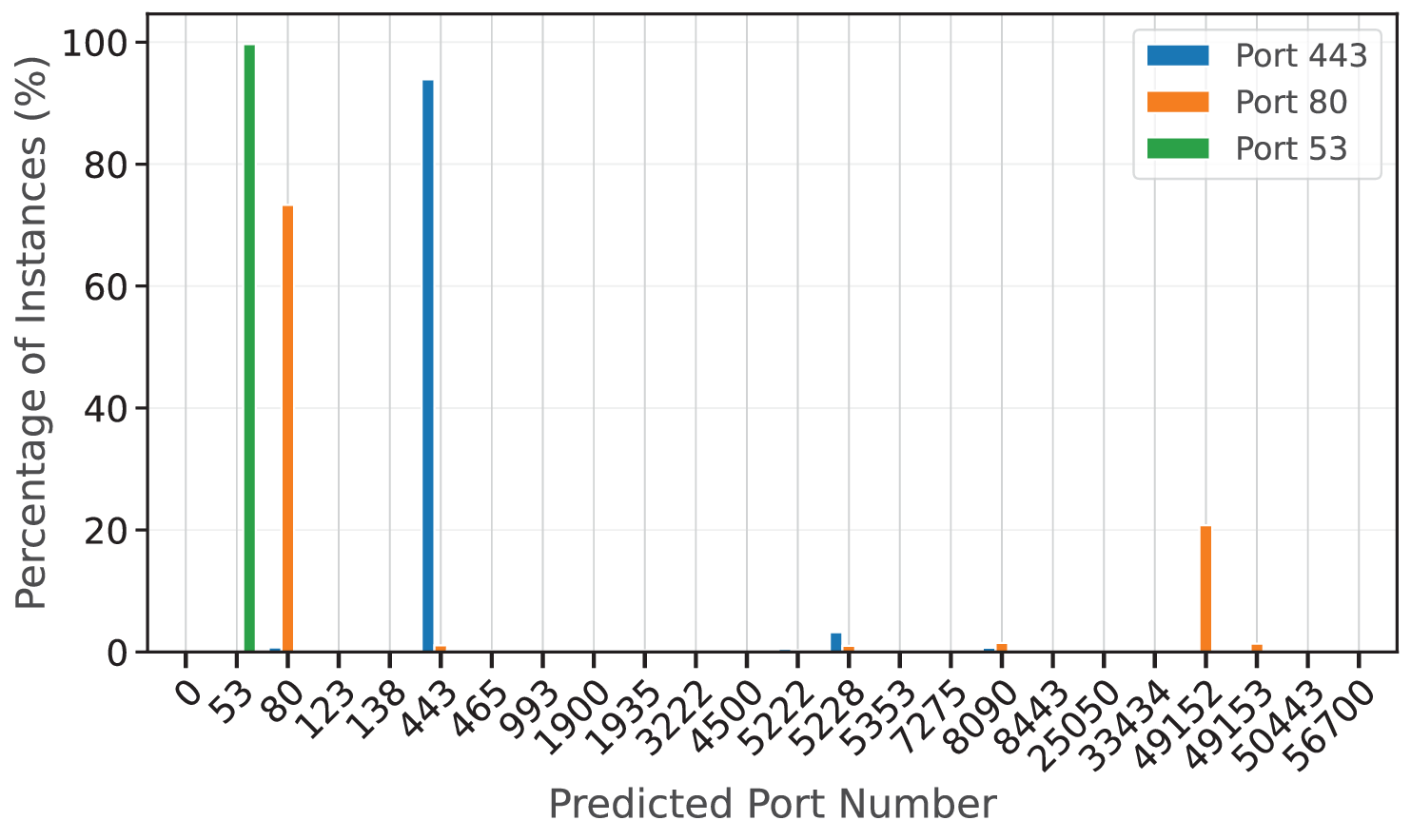}
    \caption{Distribution of reconstructed Remote Port values for three common services (ports {\myverb{443}}, {\myverb{80}}, and {\myverb{53}}) using entity embeddings.}
    \label{fig:remport_reconstruction_entity_embedding}
    \vspace{-5mm}
\end{figure}

Fig.~\ref{fig:remport_reconstruction_entity_embedding} shows the distribution of reconstructed port numbers for three well-known remote ports: {\myverb{443}} (for TLS), {\myverb{80}} (for HTTP), and {\myverb{53}} (for DNS). The model reconstructs port {\myverb{53}} with $99$\% accuracy (green bar), port {\myverb{443}} with $94$\% (blue bars), and port {\myverb{80}} with $73$\% (orange bars). In particular, $21$\% of port {\myverb{80}} instances are reconstructed as port {\myverb{49152}}, a port used by Belkin Motion and Belkin Switch devices to expose device information. Manual inspection of packet traces reveals that this service also operates over HTTP, suggesting that the reconstructed port values are influenced not only by the input port number but also by payload-level features. This indicates that the autoencoder model learns dependencies between protocol, port usage, and content-level information rather than memorizing categorical identifiers. 

Although the entity-embedding variant reconstructs categorical features effectively and maintains comparable loss for numerical attributes, it introduces additional architectural complexity. The baseline encoder \EncoderAE~ contains approximately 15M trainable parameters, whereas $\EncoderAEEntity$ increases model size to roughly 31M parameters due to the inclusion of embedding layers and a multi-branch decoder. 
We retain this variant for comparative evaluation in downstream classification and for the remainder of this section, we adopt the all–numerical–feature autoencoder as the baseline.

\subsection{Generalizability of Traffic Representations}\label{sec:vae-arch}
While autoencoders effectively minimize reconstruction error on training data, they may overfit to environment-specific patterns. VAEs address this limitation by introducing a probabilistic latent space that promotes smoother latent representations~\cite{VAE2014}.

We extend the deterministic autoencoder by parametrizing each latent dimension with a mean and standard deviation and sampling latent vectors during training. The VAE objective comprises two components: (i) a reconstruction loss, identical to that used for the autoencoder, and (ii) a Kullback--Leibler (KL) divergence term that regularizes the latent distribution toward a unit Gaussian. This regularization promotes continuity in the latent space and improves tolerance to distribution shifts.
The total loss is defined as:

\begin{equation}
    L_{\text{VAE}} = L_{\text{Reconstruction}} + \beta \cdot D_{\text{KL}}
\end{equation}
where $D_{\text{KL}}$ denotes the KL divergence and $\beta$ controls the trade-off between reconstruction fidelity and latent-space regularization. The resulting $\EncoderVAE$ contains approximately 16M trainable parameters, comparable to the 15M parameters of the baseline $\EncoderAE$.

\begin{figure}[t!]
    \centering
    \includegraphics[width=0.90\linewidth]{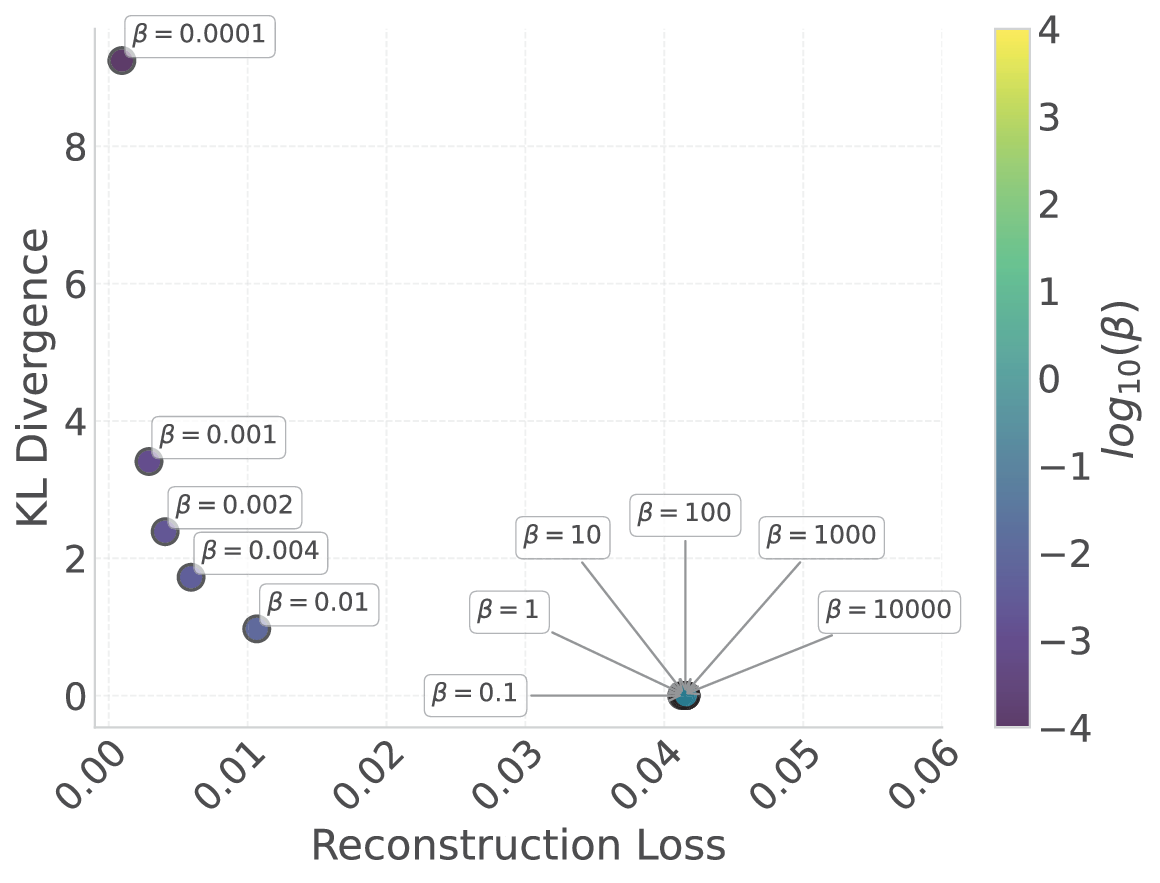}
    \caption{Trade-off between reconstruction loss and KL divergence for different values of the regularization parameter $\beta$.}
    \label{fig:vae_beta_tradeoff}
    \vspace{-5mm}
\end{figure}

We tune $\beta$ via grid search on the validation set. Fig.~\ref{fig:vae_beta_tradeoff} illustrates the trade-off between reconstruction loss and KL divergence for $\beta \in \{10^{-4}, 10^{-3}, 0.002, 0.004, 10^{-2}, 10^{-1}, 10^{0}, 10^{1}, 10^{2}, 10^{3}, 10^{4}\}$ on the validation set. As expected, increasing $\beta$ reduces KL divergence at the cost of higher reconstruction error. 
We select $\beta=0.001$ as a balanced setting that maintains low reconstruction loss while enforcing moderate latent regularization.

\begin{figure*}[t!]
    \centering
    \subfloat[Seen: trained and evaluated on full traffic distribution.]{
        \includegraphics[width=0.95\columnwidth]{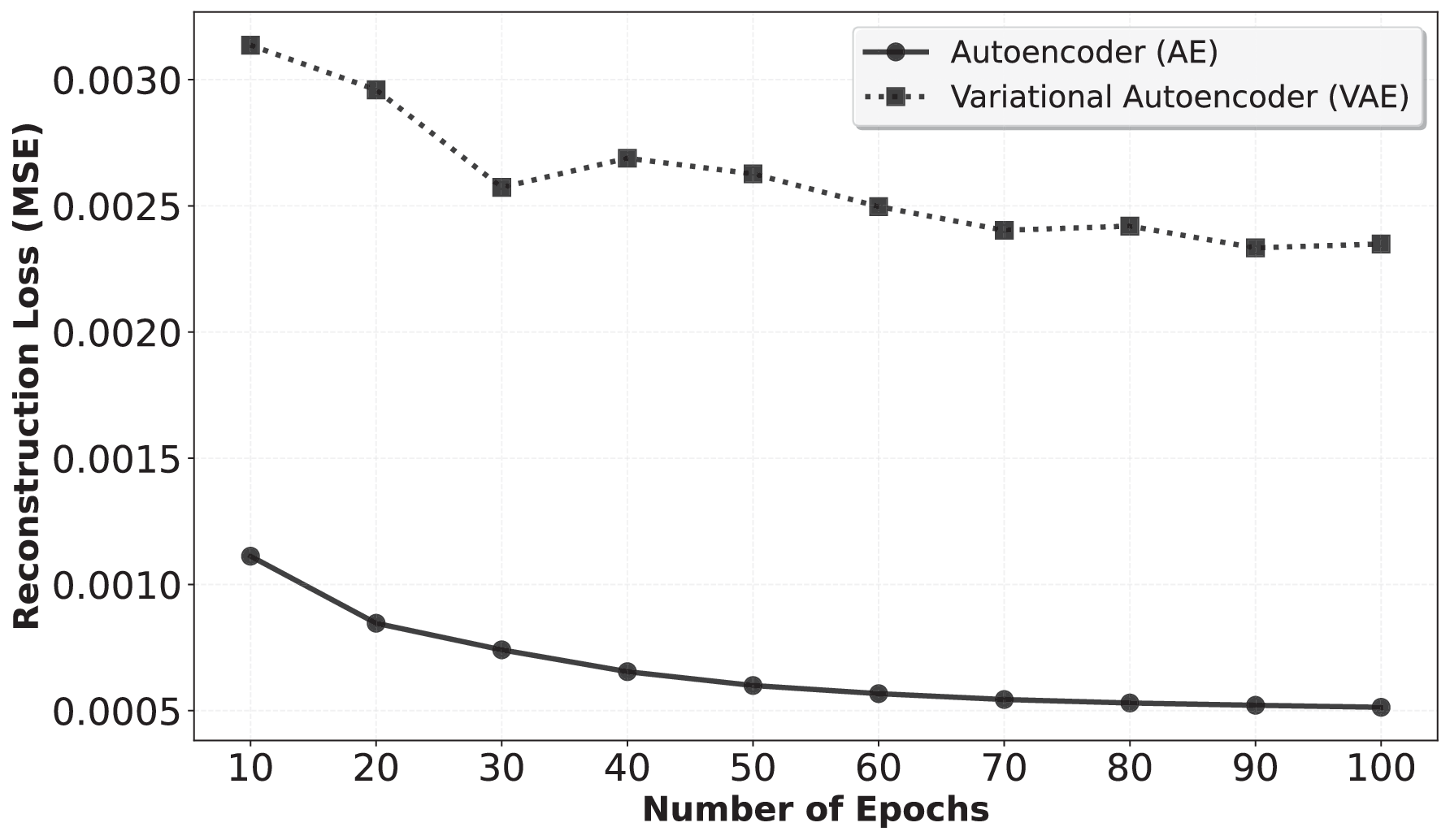}
        \label{fig:reconstruction_loss_comparison_all}
    }
    \hspace{1mm}
    \subfloat[Unseen: trained on TCP/443 only and evaluated with TCP/443 removed.]{
        \includegraphics[width=0.95\columnwidth]{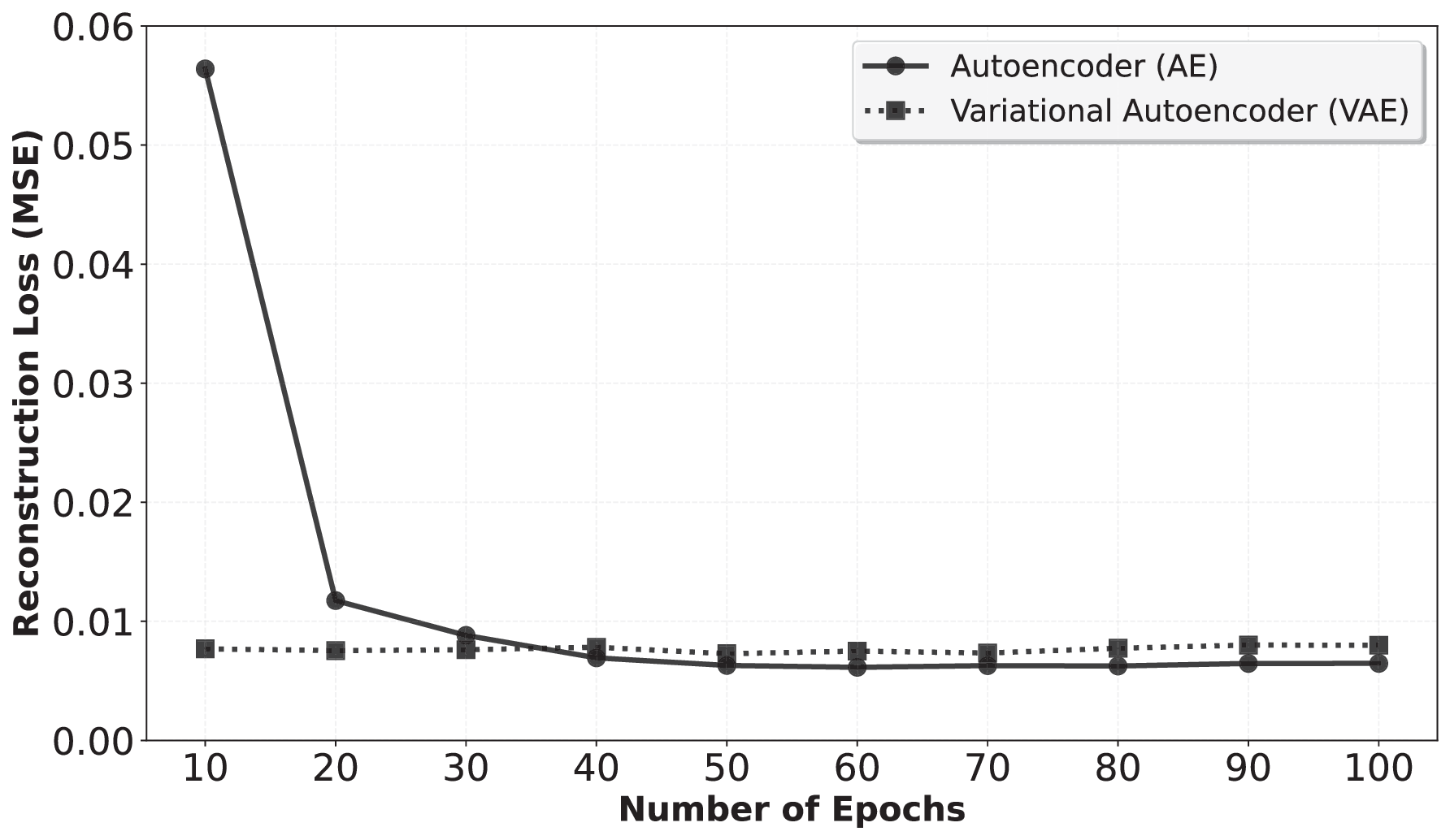}
        \label{fig:reconstruction_loss_comparison_443}
    }
    \vspace{-1mm}
    \caption{Comparison of reconstruction loss between AE and VAE models under (a) seen traffic conditions and (b) unseen traffic with distribution shift (training on TCP/443 only, evaluation with TCP/443 removed). The AE degrades sharply under distribution shift, whereas the VAE shows a much smaller increase in error, indicating greater robustness}\label{fig:reconstruction_loss_comparison}
    \vspace{-5mm}
\end{figure*}

Fig.~\ref{fig:reconstruction_loss_comparison_all} compares the reconstruction performance of $\EncoderAE$~and $\EncoderVAE$~over 100 epochs. On $\Dataset{DATA16}{val}$, the stabilized MSE of 
$\EncoderVAE$~is approximately $5.5$ times that of 
$\EncoderAE$~(\ie $0.0036$ versus $0.00065$), reflecting the expected trade-off introduced by probabilistic regularization. 

To assess robustness to unseen traffic patterns (\ie generalizability), we conduct a controlled experiment that introduces an explicit distribution shift.
In this experiment, the encoder models are trained exclusively on {\myverb{TCP/443}} flows and evaluated on validation data from which {\myverb{TCP/443}} traffic is removed. We denote these restricted-training variants as $\EncoderAELimited$ and $\EncoderVAELimited$, which use the same architectures as $\EncoderAE$ and $\EncoderVAE$ but are trained with limited traffic visibility.

We focus on {\myverb{TCP/443}} traffic as it is widely used across a broad range of IoT device types for encrypted device-to-cloud communication, making it a representative traffic class in real deployments~\cite{25TNSE}. 
This setting emulates a realistic deployment scenario in which dominant traffic classes observed during training may be absent or altered at inference time. As shown in Fig.~\ref{fig:reconstruction_loss_comparison_443}, when moving from seen traffic to the unseen traffic setting (distribution shift), the reconstruction error of the VAE model ($\EncoderVAELimited$) increases slightly to $0.007$, whereas the error for the AE model ($\EncoderAELimited$) rises sharply to $0.006$, representing nearly 10$\times$ degradation relative to its reconstruction error when trained and evaluated on the full traffic distribution. This result indicates that, despite higher reconstruction error on 
the 
seen data, the VAE-based model exhibits substantially greater robustness under distribution shift.

We emphasize that reconstruction robustness is used here as a diagnostic for latent regularization rather than as an end goal. In particular, the contribution is not the use of a variational autoencoder itself, but the empirical observation that KL-based regularization suppresses environment-specific artifacts in traffic representations and improves robustness under deployment shift. Downstream classification performance is evaluated in the next section. 
We will publicly release the implementations of the five encoder models studied in this work ($\EncoderAE$, $\EncoderVAE$, $\EncoderAEEntity$, $\EncoderAELimited$, and $\EncoderVAELimited$) at \cite{github26} 
to facilitate reproducibility and further research.


\section{IoT Device-Type Classification Using$\break$Learned Traffic Embeddings}
\label{sec:downstream-class}

With encoder models trained in the previous section, we now evaluate how their learned representations support downstream classification of IoT device types. Specifically, each encoder generates an embedding for each Custom Flow, which serves as input to a simple supervised classifier. We consider five variants of the encoder introduced earlier: $\EncoderAE$, $\EncoderVAE$, $\EncoderAEEntity$, $\EncoderVAELimited$, and $\EncoderAELimited$. Throughout this section, all encoders remain frozen to isolate the effect of representation quality from classifier capacity. Encoder architectures and model sizes are fixed as described in \S\ref{sec:traffic-representation}. In this section, we focus exclusively on the effect of representation quality on downstream classification. Accordingly, the performance differences observed in this section reflect the quality of the learned representations rather than differences in classifier capacity or training procedure. This design reflects a practical deployment model in which computationally intensive, label-free representation learning is performed once. At the same time, simple classifiers can be trained efficiently on labeled embeddings (even on CPU-only systems), highlighting that most discriminative power resides in the learned traffic representations. To assess generalizability, we evaluate the performance of the classifier in previously unseen IoT traffic from two datasets: $\Dataset{DATA16}{test}$ and $\Dataset{DATA25v1}{test}$.

\noindent \textbf{Classifier input:}
Our prior work~\cite{25TNSE} showed that classifying short batches of five past flows yields better device-level discrimination than classifying isolated flows, because the ordered sequence of flows provides additional context about device behavior.
Following this approach, we construct classifier instances by concatenating the embeddings of five successive flows observed for the same device, with each instance labeled by the corresponding device type. Each flow embedding has dimensionality 40 (as selected in \S\ref{sec:ae-hparam}), resulting in classifier input vectors of size $5 \times 40$.

\subsection{Ablation Studies}
The downstream classifier adopts a compact architecture designed to emphasize the quality of the learned embeddings rather than the complexity of the classifier. It consists of an input layer, a dropout layer, a single dense layer with $n$ units, and a softmax output layer. The output is a probability distribution over device classes. Training uses sparse categorical cross-entropy loss and the Adam optimizer~\cite{adam}. Performance is measured using the macro-averaged F1-score, which provides a balanced assessment across all device classes and is more appropriate than accuracy in multi-class settings. 

For these experiments, classifiers are trained on \Dataset{DATA16}{train} and validated on \Dataset{DATA16}{val}. Early stopping is applied when validation accuracy does not improve for five consecutive epochs, and model weights are restored from the best-performing epoch to prevent overfitting.

To identify an effective classifier configuration, we perform three ablation studies that examine key architectural and training parameters. Each experiment is repeated five times, and we report the average number of training epochs, training time, and macro F1-score.
All ablation experiments are conducted on a GPU-based server equipped with four NVIDIA L4 GPUs, each with 24 GB of GDDR6 memory.

\begin{table}[t!]
\centering
\caption{Node-count ablation study for the downstream classifier, reporting the number of trainable parameters, convergence epochs, training time, and macro F1-score (mean over five runs).}
\vspace{-2mm}
\small
\color{black}
\resizebox{0.85\columnwidth}{!}{%
\renewcommand{\arraystretch}{1.2}
\begin{tabular}{|l|r|r|r|r|}
\hline
\textbf{Nodes} &
\rotatebox{90}{\makecell{\shortstack{\textbf{$\Delta$ Params}\\ (\textit{relative, \%})}}} &
\rotatebox{90}{\makecell{\textbf{Epochs} \\ (\textit{early-stop})}} &
\rotatebox{90}{\makecell{\shortstack{\textbf{$\Delta$ Train Time}\\ (\textit{relative, \%})}}} &
\rotatebox{90}{\makecell{\shortstack{\textbf{$\Delta$ F1-score} \\ (\textit{relative, \%})}}} \\
\hline
40 (\textit{ref}) & \textbf{$9$K} & \textbf{$83$} & \textbf{$65$ s} & \textbf{$0.9490$} \\ \hline
64 & +$60$\% & $70$ & -$16$\% & +$0.42$\% \\\hline
128 & +$220$\% & $61$ & -$26$\% & +$1$\% \\\hline
256 & +$539$\% & $56$ & -$31$\% & +$2$\% \\\hline
512 & +$1177$\% & $41$ & -$47$\% & +$2$\% \\
\hline
\end{tabular}
\label{tab:node_ablation}
\vspace{-7mm}
}
\end{table}

\vspace{1mm}
\noindent\textbf{Node-Count Ablation:} This study evaluates the impact of dense-layer width by testing configurations with 40, 64, 128, 256, and 512 units (40 units correspond to the per-flow embedding dimensionality). Table~\ref{tab:node_ablation} summarizes the resulting number of trainable parameters, average convergence epochs, average training time, and macro F1-score (each value is the mean over five runs). 
As expected, larger layers increase the parameter count. Interestingly, larger dense layers reduce the number of epochs to convergence and total training time on our GPU: (i) increased expressiveness accelerates optimization, and (ii) larger matrix operations improve GPU utilization and throughput. However, the macro F1-score improvement saturates beyond 256 units, even though the training time decreases by approximately $31$\% compared to 40 units. Although increasing the number of units from 256 to 512 yields a modest further reduction in training time, it yields negligible improvement in F1 and increases the risk of overfitting; dropout mitigates, but does not eliminate this risk. Given the marginal gains beyond 256 units, we adopt 256 units for the remaining experiments. In this setting, the downstream classifier has approximately 56K trainable parameters, which we use as the reference configuration for the subsequent ablation experiment.

\begin{table}[t!]
\centering
\caption{Layer-depth ablation study for the downstream classifier, reporting the number of trainable parameters, convergence epochs, training time, and macro F1-score (mean over five runs).}
\vspace{-2mm}
\small
\color{black}
\resizebox{0.85\columnwidth}{!}{%
\renewcommand{\arraystretch}{1.2}
\begin{tabular}{|l|r|r|r|r|}
\hline
\textbf{Dense Blocks} &
\rotatebox{90}{\makecell{\shortstack{\textbf{$\Delta$ Params}\\ (\textit{relative, \%})}}} &
\rotatebox{90}{\makecell{\textbf{Epochs}}} &
\rotatebox{90}{\makecell{\shortstack{\textbf{$\Delta$ Train Time}\\ (\textit{relative, \%})}}} &
\rotatebox{90}{\makecell{\shortstack{\textbf{$\Delta$ F1-score} \\ (\textit{relative, \%})}}} \\
\hline
1 (\textit{ref}) & \textbf{$56$K} & \textbf{$55$} & \textbf{$44$ s} & \textbf{$0.9631$} \\\hline
2 & +$54$\% & $45$ & -$0.08$\% & +$1$\% \\\hline
3 & +$67$\% & $51$ & +$45$\% & +$1$\% \\\hline
4 & +$69$\% & $54$ & +$71$\% & +$1$\% \\
\hline
\end{tabular}
\label{tab:layer_ablation}
\vspace{-6mm}
}
\end{table}

\vspace{1mm}
\noindent
\textbf{Layer-Depth Ablation:} 
Next, using the 256-unit reference configuration, we investigate whether increasing the classifier depth significantly improves performance.
We evaluate architectures composed of one to four dropout--dense blocks, where each successive dense layers halve the number of units (256, 128, 64, 32). Each block comprises a dropout layer with a rate of 0.3, followed by a dense layer. 
The 
results in Table~\ref{tab:layer_ablation} show that deeper configurations substantially increase training time and convergence epochs (up to $71$\% longer), while producing only marginal improvements in macro F1-score. Given the substantially higher computational cost and limited accuracy gains, a single dropout--dense block (256 units) is sufficient for our downstream evaluation.
As a result, the selected classifier configuration maintains a model size of approximately 56K trainable parameters.

\vspace{1mm}
\noindent
\textbf{Maximum Instances per Class:} Finally, we investigate the effect of training-set size by capping the maximum number of instances per class. We experiment with caps ranging from 125 to 8,000 instances per class, increasing the total training set from roughly 2.3k to 38k samples across 19 classes (some classes have fewer samples and do not reach the cap). 
Table~\ref{tab:max_instances_ablation} reports the average convergence epochs, average training time, and macro F1-score (means over five runs).
As the instance cap increases, the number of epochs required for convergence decreases overall (from $86$ to $51$), while total training time increases (from $42$\,s up to about $75$\,s), reflecting the typical trade-off between iteration count and per-epoch cost. 
Macro F1 improves substantially when increasing the cap from 250 to 1,000 instances per class; gains beyond 1,000 instances are marginal. Based on this observation, we fix the maximum number of training instances per class to 1,000 for subsequent experiments.
All experiments use a batch size of 16 and a learning rate of 0.01. 

\begin{table}[t!]
\centering
\caption{Training-data ablation study for the downstream classifier, showing the effect of increasing the maximum number of training instances per class on convergence behavior, training time, and macro F1-score (mean over five runs).}
\vspace{-2mm}
\small
\color{black}
\resizebox{0.85\columnwidth}{!}{%
\renewcommand{\arraystretch}{1.2}
\begin{tabular}{|l|r|r|r|r|}
\hline
\textbf{Max Instances} &
\rotatebox{90}{\makecell{\textbf{Train Samples}}} &
\rotatebox{90}{\makecell{\textbf{Epochs} \\ (\textit{early-stop})}} &
\rotatebox{90}{\makecell{\shortstack{\textbf{$\Delta$ Train Time}\\ (\textit{relative, \%})}}} &
\rotatebox{90}{\makecell{\shortstack{\textbf{$\Delta$ F1-score} \\ (\textit{relative, \%})}}} \\
\hline
125 (\textit{ref}) & \textbf{2k} & \textbf{$86$} & \textbf{$42$ s} & \textbf{$0.9110$} \\\hline
250 & 5k & $65$ & -$15$\% & +$3$\% \\\hline
500 & 10k & $55$ & -$16$\% & +$5$\% \\\hline
1k & 19k & $60$ & +$11$\% & +$7$\% \\\hline
2k & 38k & $46$ & +$24$\% & +$8$\% \\\hline
4k & 38k & $65$ & +$75$\% & +$8$\% \\\hline
8k & 38k & $51$ & +$42$\% & +$8$\% \\
\hline
\end{tabular}
\label{tab:max_instances_ablation}
\vspace{-7mm}
}
\end{table}

\subsection{Model Performance}
We now evaluate the performance of the downstream classifier using embeddings generated by the five encoders introduced earlier: 
$\EncoderAE$, $\EncoderVAE$, $\EncoderAEEntity$, $\EncoderAELimited$, and $\EncoderVAELimited$. For all experiments, classifiers are trained using  $\Dataset{DATA16}{train}$ with the optimal hyperparameters identified in ablation studies: dense layer size of 256, one dropout--dense block, a maximum of 1,000 instances per class, learning rate of 0.01, and batch size of 64.

\begin{figure}[!t]
    \centering
    \includegraphics[width=1\linewidth]{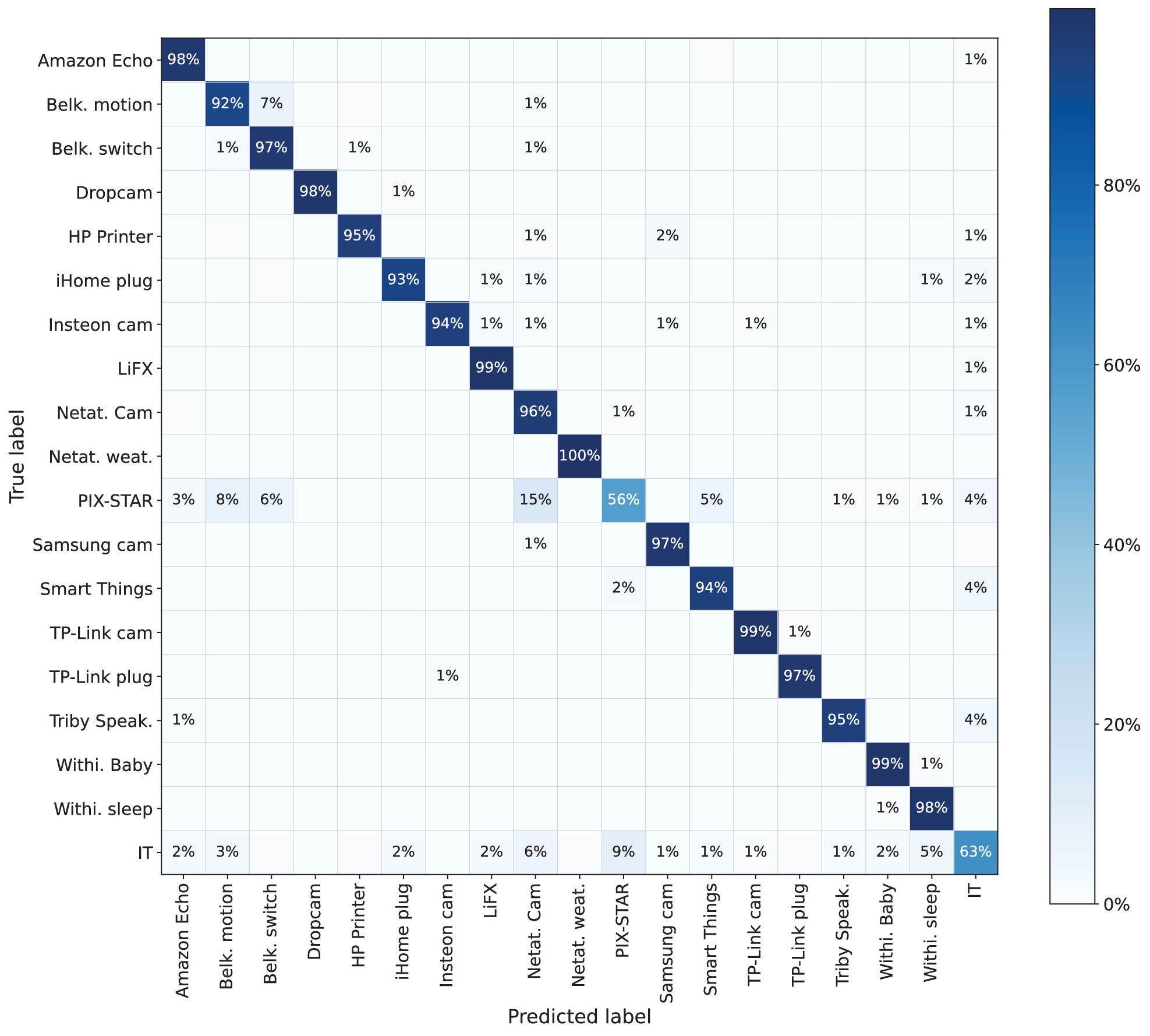}
    \caption{Confusion matrix of the downstream classifier trained on embeddings generated by $\EncoderAE$, evaluated on \Dataset{DATA16}{test}.}
    \label{fig:cm-ae}
    \vspace{-5mm}
\end{figure}

\begin{figure*}[t]
    \centering
    \subfloat[$\EncoderAELimited$.]{
        \includegraphics[width=0.47\textwidth]{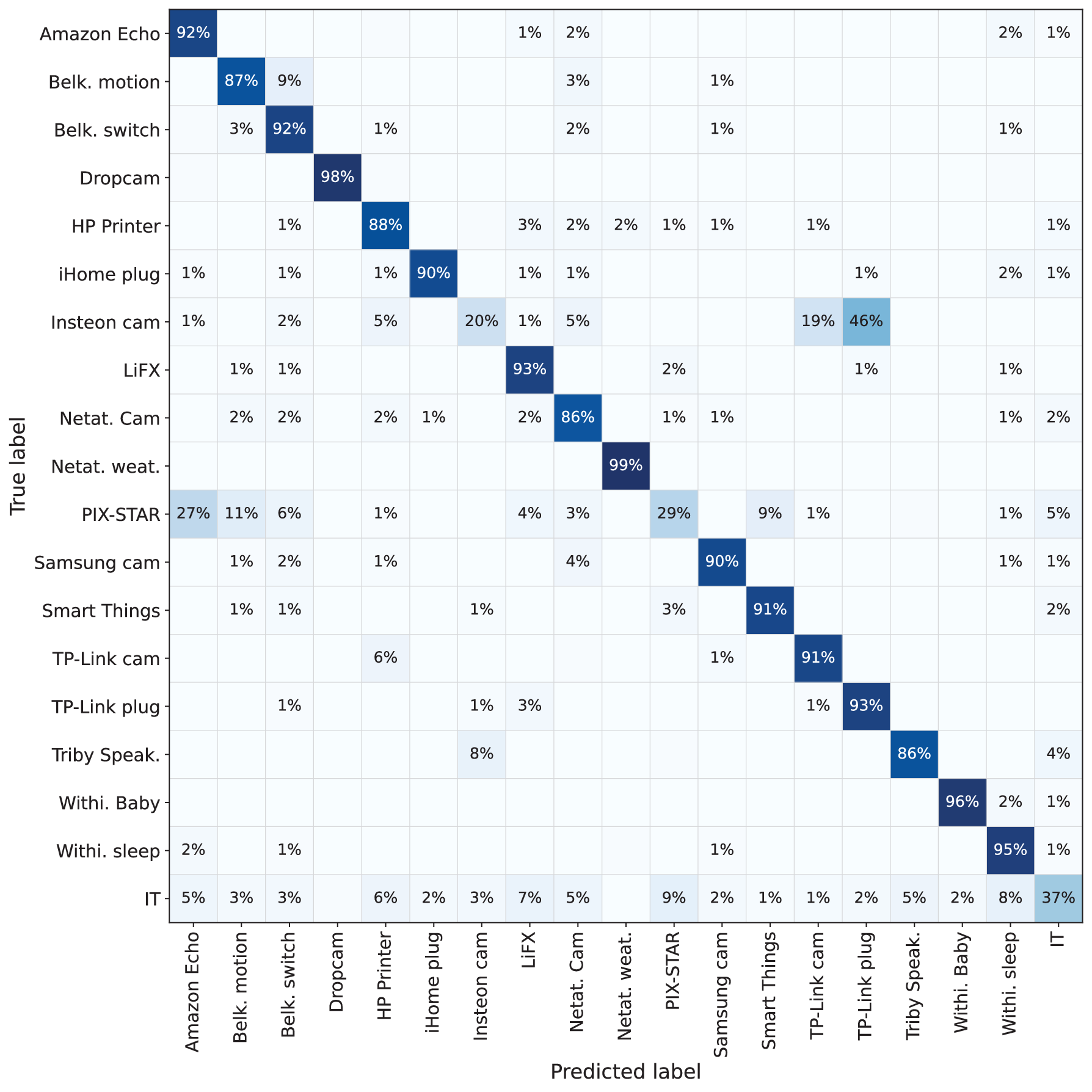}
        \label{fig:cm-ae-limited}
    }
    \hfill
    \subfloat[$\EncoderVAELimited$.]{
        \includegraphics[width=0.47\textwidth]{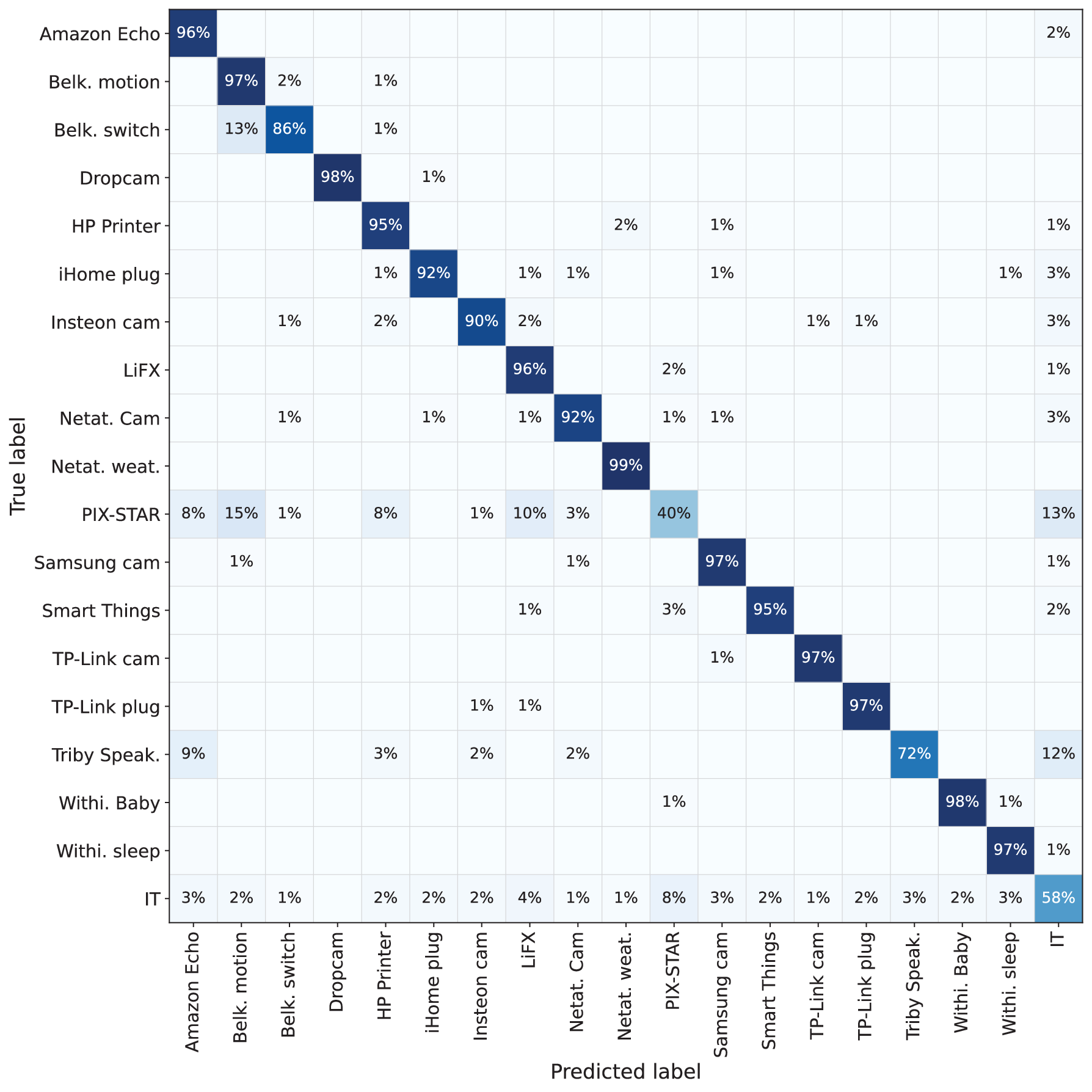}
        \label{fig:cm-vae-limited}
    }
    \caption{Confusion matrices of downstream classifiers trained on embeddings generated by (a) $\EncoderAELimited$ and (b) $\EncoderVAELimited$, evaluated on $\Dataset{DATA16}{test}$.}
    \label{fig:cm-limited}
    \vspace{-6mm}
\end{figure*}

We first train separate classifiers for embeddings produced by each encoder using $\Dataset{DATA16}{train}$ and validate using $\Dataset{DATA16}{val}$. The classification performance is then evaluated on randomly sampled instances from $\Dataset{DATA16}{test}$, with a cap of 10,000 instances per class.

Fig.~\ref{fig:cm-ae} shows the confusion matrix for the classifier trained on \EncoderAE~embeddings. The matrix exhibits a near-diagonal structure, indicating strong per-class performance. Several device types, such as Netatmo Weather, LiFX, Samsung Cam, and TP-Link Cam, achieve perfect or near-perfect classification ($\geq 99\%$), while 12 of the 19 classes exceed $95$\% accuracy. However, PIX-STAR and \textit{IT} devices exhibit lower performance, with accuracies of $56$\% and $68$\%, respectively. Our previous work~\cite{25TNSE} reported similar challenges for PIX-STAR photo frames, which lack distinctive traffic fingerprints compared to devices with more consistent behavioral patterns (\eg Amazon Echo or Netatmo Camera). In addition, a significant portion of the PIX-STAR test instances involve photo uploads via HTTPS, a behavior not observed during training. 

The \textit{IT} class aggregates multiple non-IoT devices (\eg laptops, tablets, and mobile phones). Unlike IoT devices, the traffic behavior of \textit{IT} devices is highly influenced by user activity, making it difficult to capture all relevant patterns during training. Furthermore, limiting the maximum number of training instances per class reduces behavioral diversity. When the instances cap is increased from 1,000 to 8,000, the classification accuracy for the \textit{IT} class improves from $63$\% to $83$\%. This observation highlights the sensitivity of highly variable and aggregated classes to training data volume, in contrast to device-specific IoT classes, which tend to exhibit stable traffic patterns.

We observe similar confusion matrices for classifiers trained on embeddings generated by $\EncoderVAE$ and $\EncoderAEEntity$ (figures omitted due to space constraints). The overall classification accuracies for $\textrm{AE}$, $\textrm{VAE}$, and $\textrm{AE}_{\text{entity}}$ are $94$\%, $93$\%, and $95$\%, respectively, with corresponding macro F1-scores of $0.95$, $0.93$, and $0.94$. It is important to emphasize that $\textrm{AE}$ and $\textrm{VAE}$ treat all features as numerical, whereas $\textrm{AE}_{\text{entity}}$ explicitly models Protocol, Device Port, and Remote Port as categorical features. Note that all three encoders are trained on the same full \Dataset{DATA16}{train} dataset.

Fig.~\ref{fig:cm-limited} presents the confusion matrices for classifiers trained on embeddings from $\EncoderAELimited$ and $\EncoderVAELimited$. Both models show reduced performance relative to their full-data counterparts. Specifically, $\textrm{AE}_{\text{limited}}$ achieves an overall accuracy of $83$\% with a macro F1-score of $0.80$, while $\textrm{VAE}_{\text{limited}}$ attains $91$\% accuracy and a macro F1-score of $0.89$.

Overall, these results indicate that $\textrm{AE}$, $\textrm{VAE}$, and $\textrm{AE}_{\text{entity}}$ achieve comparable performance when classifying device types within the same traffic domain. However, when encoders are trained on restricted traffic conditions, representation quality degrades significantly for $\textrm{AE}$, whereas $\textrm{VAE}$ maintains consistent performance, reflecting its improved robustness to distribution shift. These results confirm that, when trained and evaluated within the same traffic domain, the choice of representation dominates downstream classification performance.

\begin{table*}[b]
\centering
\caption{Overall classification accuracy and macro F1-score of downstream classifiers using different encoder-generated embeddings, evaluated on \Dataset{DATA16}{} and \Dataset{DATA25v1}{}.}
\label{tab:model_comparison}
\begin{tabular}{lcccccccccc}
\toprule
 & \multicolumn{2}{c}{$\EncoderAE$} & \multicolumn{2}{c}{$\EncoderVAE$} & \multicolumn{2}{c}{$\EncoderAEEntity$} & \multicolumn{2}{c}{$\EncoderAELimited$} & \multicolumn{2}{c}{$\EncoderVAELimited$} \\
\cmidrule(lr){2-3} \cmidrule(lr){4-5} \cmidrule(lr){6-7} \cmidrule(lr){8-9} \cmidrule(lr){10-11}
Environment & Acc(\%) & F1 & Acc(\%) & F1 & Acc(\%) & F1 & Acc(\%) & F1 & Acc(\%) & F1 \\
\midrule
$\Dataset{DATA16}{}$ & $94$ & $0.93$ & $93$ & $0.92$ & $94$ & $0.93$ & $83$ & $0.80$ & $91$ & $0.89$ \\
$\Dataset{DATA25_{v1}}{}$ & $92$ & $0.91$ & $89$ & $0.88$ & $91$ & $0.91$ & $78$ & $0.77$ & $84$ & $0.83$ \\
\bottomrule
\end{tabular}
\end{table*}

\subsection{Generalizability of Encoders for Classification}
We evaluate the generalizability of the pretrained encoders by applying them to a second dataset, \Dataset{DATA25v1}{}, and training classifiers on 19 device classes whose traffic 
is 
not observed during encoder training. Since all encoders are pretrained exclusively on \Dataset{DATA16}{}, the traffic in \Dataset{DATA25v1}{} is completely unseen during representation learning.

\begin{figure}[!tb]
    \centering
    \includegraphics[width=1\linewidth]{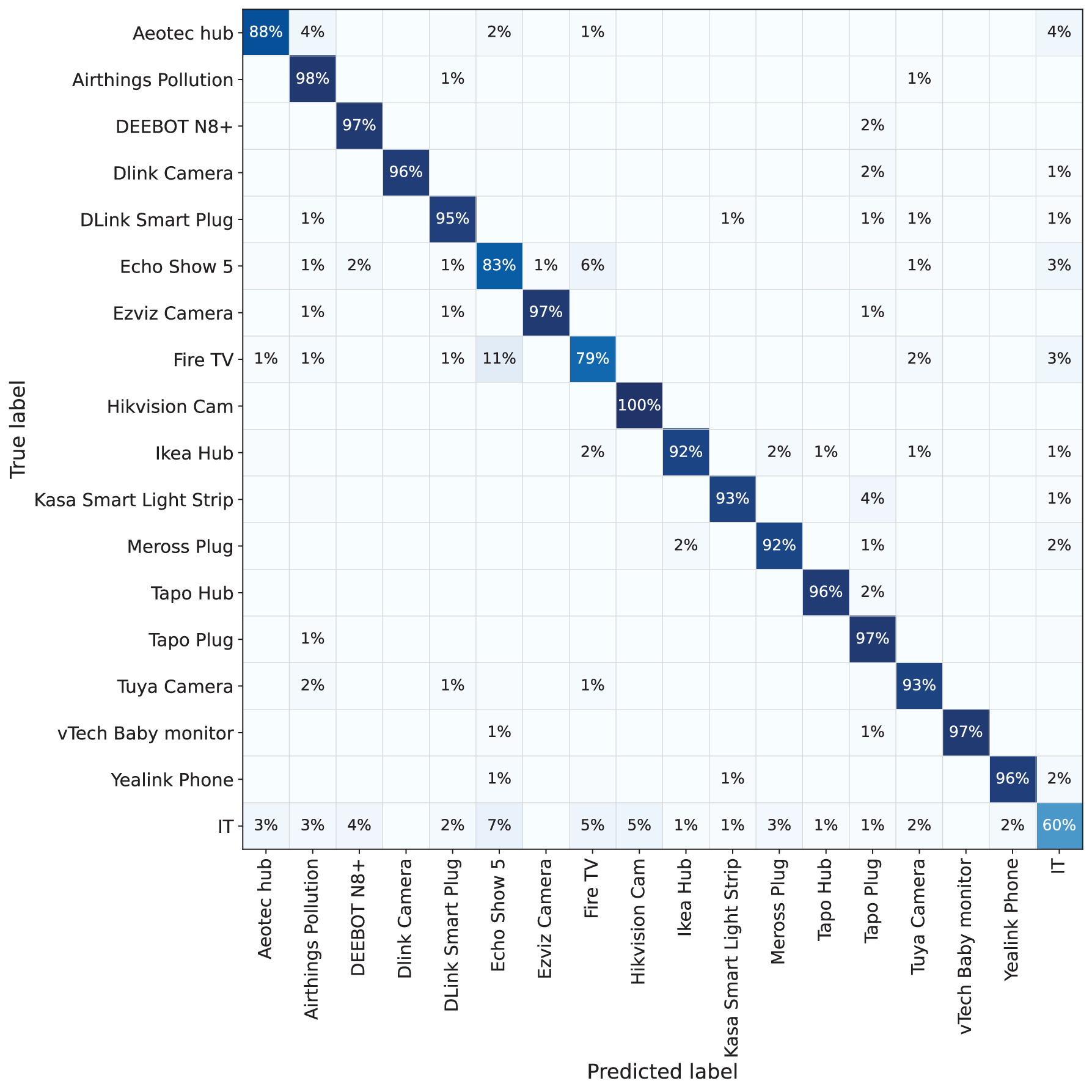}
    \caption{Confusion matrix of the downstream classifier trained on embeddings generated by $\EncoderAE$, evaluated on \Dataset{DATA25v1}{}.}
    \label{fig:cm-data25-ae}
    \vspace{-5mm}
\end{figure}

Fig.~\ref{fig:cm-data25-ae} shows the confusion matrix of the classifier trained on embeddings generated by $\EncoderAE$ for \Dataset{DATA25v1}{}. Note that device classes differ from those seen in the confusion matrices for \Dataset{DATA16}{test}. This matrix is largely diagonal, yielding a macro F1-score of $0.93$ and an overall classification accuracy of $94$\%. Nevertheless, several classes exhibit moderate performance (\ie below $90$\%). For example, $11$\% of Fire TV instances are misclassified as Echo Show, and $6$\% of Echo Show instances are misclassified as Fire TV. This behavior is expected because (i) both devices are developed by Amazon and therefore share similar traffic characteristics, and (ii) the Echo Show frequently operates as a control hub, generating regular communication with the Fire TV, which leads to overlapping behavioral patterns.

Table~\ref{tab:model_comparison} compares the overall accuracy and macro F1-score of classifiers trained on \Dataset{DATA25v1}{} with those obtained on \Dataset{DATA16}{}.
As observed, both datasets exhibit similar performance trends across different encoder-based embeddings. In both cases, $\textrm{AE}$, $\textrm{VAE}$, and $\textrm{AE}_{\text{entity}}$ achieve comparable performance, while 
$\textrm{VAE}_{\text{limited}}$ consistently outperforms 
$\textrm{AE}_{\text{limited}}$ by approximately $6-9$\%.

\section{Benchmarking the Encoder Models}\label{sec:benchmark}

In addition to comparing different encoder architectures, we explicitly evaluate cross-environment generalization by training classifiers on \Dataset{DATA25v1}{} and testing them on \Dataset{DATA25v2}{}, which contains overlapping device classes collected in a different deployment environment.

\subsection{Benchmarking Setup and Baselines}

We benchmark our encoders against two recently proposed traffic encoders, namely ET-BERT~\cite{ETBERT2022} and NetMamba~\cite{Netmamba2024}, using a unified evaluation protocol. 
In all cases, pretrained encoders are used strictly as frozen feature extractors, and the same lightweight downstream classifier described in \S\ref{sec:downstream-class} is trained on the resulting embeddings to ensure a fair comparison.



ET-BERT is a pretraining-based encoder for encrypted network traffic that employs a bidirectional Transformer architecture inspired by BERT \cite{ETBERT2022}. It models byte-level dependencies within encrypted datagram sequences and comprises 12 Transformer encoder blocks, with approximately 135 million trainable parameters.
ET-BERT groups packets into flows using standard five-tuple identifiers and removes all link-, network-, and transport-layer headers, and aggregates packets into \textit{BURSTs} of time-adjacent packets in the same direction. The preprocessing pipeline filters for TCP/UDP flows with at least five packets and does not apply flow timeouts. 



NetMamba is a pretraining-based encoder built on a linear-time state-space model architecture using the unidirectional Mamba formulation~\cite{Netmamba2024}. It consists of four Mamba blocks and contains approximately 1.89 million parameters. 
NetMamba segments traffic using five-tuple identifiers without timeouts. and extracts fixed-size representations by concatenating header and payload bytes from the first five packets of each flow, discarding subsequent packets. 

Taken together, our convolutional autoencoder-based models ($\EncoderAE$ and $\EncoderVAE$), ET-BERT, and NetMamba span a broad range of model sizes and design philosophies, from moderate-size encoder--decoder architectures to large transformer-based models and compact state-space models, enabling a comparative evaluation that isolates the effect of traffic representation design rather than classifier capacity.

To align both models with our benchmarking protocol, we remove their original classification heads and extract fixed per-flow embeddings from their internal classification tokens: the {\myverb{[CLS]}} (classification) token of dimension $1\times768$ for ET-BERT and
the internal {\myverb{[CLS]}} token of dimension $1\times256$ for NetMamba.
These embeddings are then batched into instances of five successive flows and passed to the same downstream classifier used throughout this section, with encoder weights kept frozen.


\begin{table*}[t!]
\centering
\caption{Overall classification accuracy and macro F1-score for ET-BERT and NetMamba under original and locally pretrained settings, evaluated on \Dataset{DATA16}{} and \Dataset{DATA25v1}{} using a frozen-encoder protocol.}
\label{tab:model_benchmark}
\begin{tabular}{lcccccccc}
\toprule
 & \multicolumn{4}{c}{\textbf{ET-BERT}} & \multicolumn{4}{c}{\textbf{NetMamba}} \\
\cmidrule(lr){2-5} \cmidrule(lr){6-9}
 & \multicolumn{2}{c}{Original} & \multicolumn{2}{c}{Pretrained} & \multicolumn{2}{c}{Original} & \multicolumn{2}{c}{Pretrained} \\
\cmidrule(lr){2-3} \cmidrule(lr){4-5} \cmidrule(lr){6-7} \cmidrule(lr){8-9}
\textbf{Dataset} & Acc(\%) & F1 & Acc(\%) & F1 & Acc(\%) & F1 & Acc(\%) & F1 \\
\midrule
$\Dataset{DATA16}{}$      & $58$ & $0.59$ & $70$ & $0.68$ & $39$ & $0.28$ & $56$ & $0.46$ \\
$\Dataset{DATA25_{v1}}{}$ & $79$ & $0.70$ & $85$ & $0.75$ & $25$ & $0.23$ & $41$ & $0.36$ \\
\bottomrule
\end{tabular}
\vspace{-3mm}
\end{table*}

It is important to note that the primary distinction between our flow representation and those used by ET-BERT and NetMamba arises in the presence of long-lived connections, which are common in IoT deployments. In our approach, long flows are segmented into fixed one-minute intervals, allowing a persistent connection to contribute multiple flow instances over time and enabling progressive, near-real-time inference.
In contrast, ET-BERT and NetMamba treat a long-lived connection as a single flow instance and either aggregate packets until the end of the flow or rely solely on the initial packets, which can delay inference or ignore substantial portions of sustained traffic. While all approaches impose packet limits per flow, our timeout-based segmentation repeatedly extracts information from ongoing connections, utilizing more of the available traffic and supporting timely classification in operational settings.

\subsection{Benchmarking Results}
We evaluate ET-BERT and NetMamba under the same frozen-encoder protocol. 
For each model, we evaluate two variants: (i) the original pretrained encoder released by the authors, and  (ii) a locally pretrained variant trained from scratch on \Dataset{Data16}{train}, which contains both encrypted and unencrypted IoT traffic.
In all cases, per-flow embeddings are extracted, five successive flows are batched into a single classification instance, and the same simple downstream classifier described in \S\ref{sec:downstream-class} is trained on top of the fixed embeddings. Because different encoders produce embeddings of different dimensionalities, the input layer of the classifier is adjusted accordingly, while all other architectural components and hyperparameters remain unchanged. Table~\ref{tab:model_benchmark} summarizes the performance results of downstream classification.

Across all settings, trends observed in macro F1-score are consistent with those reflected in overall accuracy, indicating stable performance across device classes. Overall, ET-BERT consistently outperforms NetMamba by a substantial margin across both datasets and training regimes, achieving higher macro F1-scores and overall accuracies.

Both encoders benefit from local pretraining, indicating that representation quality is sensitive to the underlying traffic distribution; for example, ET-BERT was originally trained on generic encrypted traffic rather than IoT-specific data \cite{ETBERT2022}. For ET-BERT, local pretraining improves macro F1-score from $0.59$ to $0.68$ on \Dataset{DATA16}{} and from $0.70$ to $0.75$ on \Dataset{DATA25v1}{}. NetMamba also benefits from local pretraining, with a macro F1-score increasing from $0.28$ to $0.46$ on $\Dataset{DATA16}{}$ and from $0.23$ to $0.36$ on \Dataset{DATA25v1}{}. These improvements demonstrate the value of domain-specific IoT traffic for representation learning, but do not eliminate the performance gap between the two encoder architectures.

Despite the benefits of local pretraining, neither ET-BERT nor NetMamba achieves downstream classification performance comparable to that of our encoders under the same frozen-embedding protocol. As shown in Table~\ref{tab:model_benchmark}, the best-performing ET-BERT configuration attains a macro F1-score of $0.68$ on \Dataset{DATA16}{} and $0.75$ on \Dataset{DATA25v1}{}, while NetMamba reaches only $0.46$ and $0.36$, respectively. In contrast, our encoders consistently achieve macro F1-scores above $0.9$ on both datasets (see Table~\ref{tab:model_comparison}). This persistent gap indicates that differences in traffic representation and flow handling, rather than classifier capacity or pretraining alone, play a dominant role.

\begin{figure*}[t!]
    \centering
    \subfloat[Original ET-BERT.]{
        \includegraphics[width=0.47\textwidth]{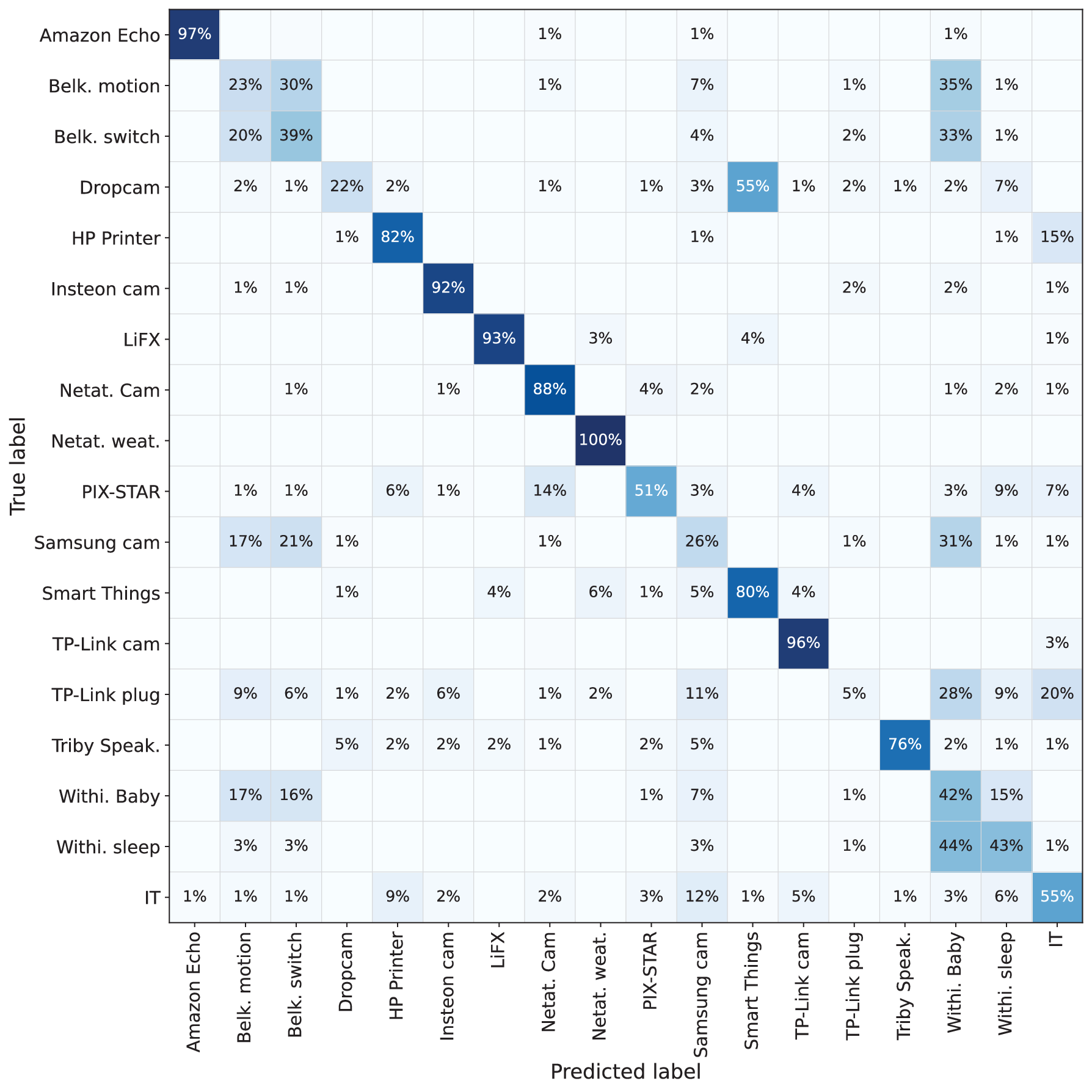}
        \label{fig:ETBERT-orig}
    }
    \hfill
    \subfloat[Locally pretrained ET-BERT.]{
        \includegraphics[width=0.47\textwidth]{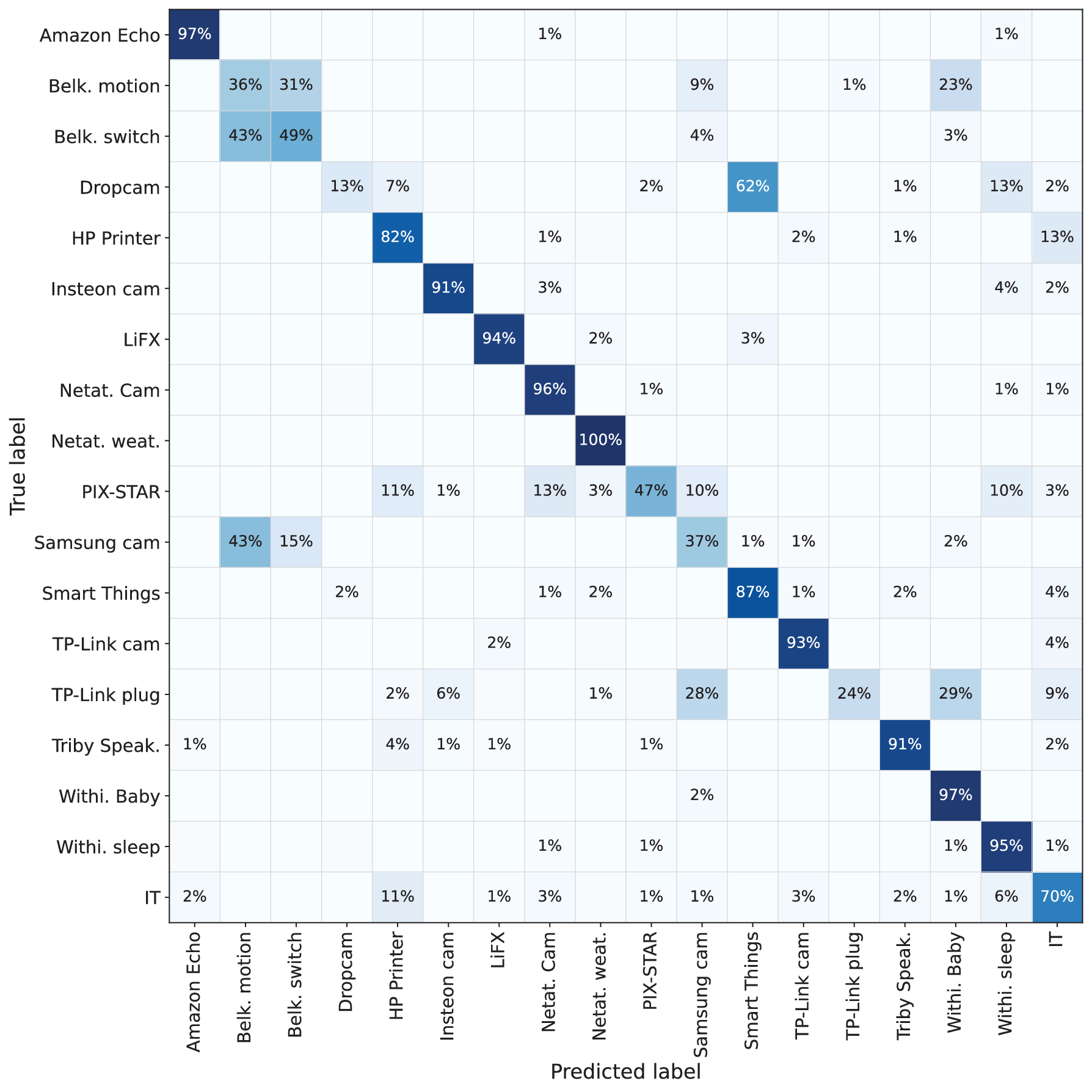}
        \label{fig:ETBERT-pt}
    }
    \caption{Confusion matrices of downstream classifiers trained on embeddings generated by: (a) original ET-BERT, and (b) locally pretrained ET-BERT, evaluated on $\Dataset{DATA16}{test}$.}
    \label{fig:ETBERT}
    \vspace{-5mm}
\end{figure*}

To better understand the impact of local pretraining at the class level, we examine the confusion matrices of ET-BERT before and after local pretraining (see Fig.~\ref{fig:ETBERT}). Local pretraining yields a more diagonal confusion matrix, indicating improved recall across several device classes. In particular, classes such as Withings Baby Monitor, Withings Sleep Sensor, Triby Speaker, and \textit{IT} show substantial recall gains (by up to $50$\%), relative to the original ET-BERT model. However, several challenging classes remain difficult even after local pretraining, including Belkin Switch, Belkin Motion, Dropcam, and TP-Link Plug.

Compared with the confusion matrices obtained using our encoders (Fig.~\ref{fig:cm-ae} and~\ref{fig:cm-limited}), we make two observations. The PIX-STAR photo frame continues to exhibit limited recall (approximately $50$\%), reflecting its weak and highly variable traffic signature. 
The \textit{IT} class shows slightly higher recall under ET-BERT (around $70$\%) than under our encoders (approximately $63$\%), suggesting that payload-centric encoders may better capture user-driven variability in non-IoT devices.

When interpreting these benchmarking results, it is important to consider key design choices in the ET-BERT and NetMamba pipelines. ET-BERT does not apply flow timeouts, removes all protocol headers, restricts analysis to TCP/UDP flows with at least five packets, and discards short flows, eliminating most of the DNS and NTP traffic. As a result, devices with long-lived sessions (\eg Dropcam) experience a severe reduction in usable instances compared to our approach. NetMamba similarly does not apply timeouts and retains only the first five packets of each flow, discarding the remaining packets from long-lived connections, and thereby limiting information extraction.

\subsection{Cross-Environment Evaluation Results}

We now evaluate cross-environment generalizability by applying trained downstream classifiers to traffic collected in a previously unseen deployment setting. As described in \S\ref{sec:data}, \Dataset{DATA25v2}{} contains ten device types that overlap with \Dataset{DATA25v1}{}, but was captured on a different university campus network with distinct topology, background traffic, and user activity patterns. To assess robustness under such conditions, we evaluate classifiers trained on \Dataset{DATA25v1}{} using embeddings generated by \EncoderAE, \EncoderAEEntity, \EncoderVAE, \EncoderAELimited, and \EncoderVAELimited. Both the encoder and classifier remain frozen, and no fine-tuning or adaptation is performed.

Across encoders, the resulting macro F1-scores on \Dataset{DATA25v2}{} are $0.71$, $0.69$, $0.80$, $0.62$, and $0.78$ for \EncoderAE, \EncoderAEEntity, \EncoderVAE, \EncoderAELimited, and \EncoderVAELimited, respectively, with corresponding accuracies of $72$\%, $70$\%, $78$\%, $59$\%, and $75$\%. These results indicate a clear performance separation between VAE-based and AE-based representations under cross-environment deployment.

\begin{figure}[t!]
    \centering
    \includegraphics[width=\linewidth]{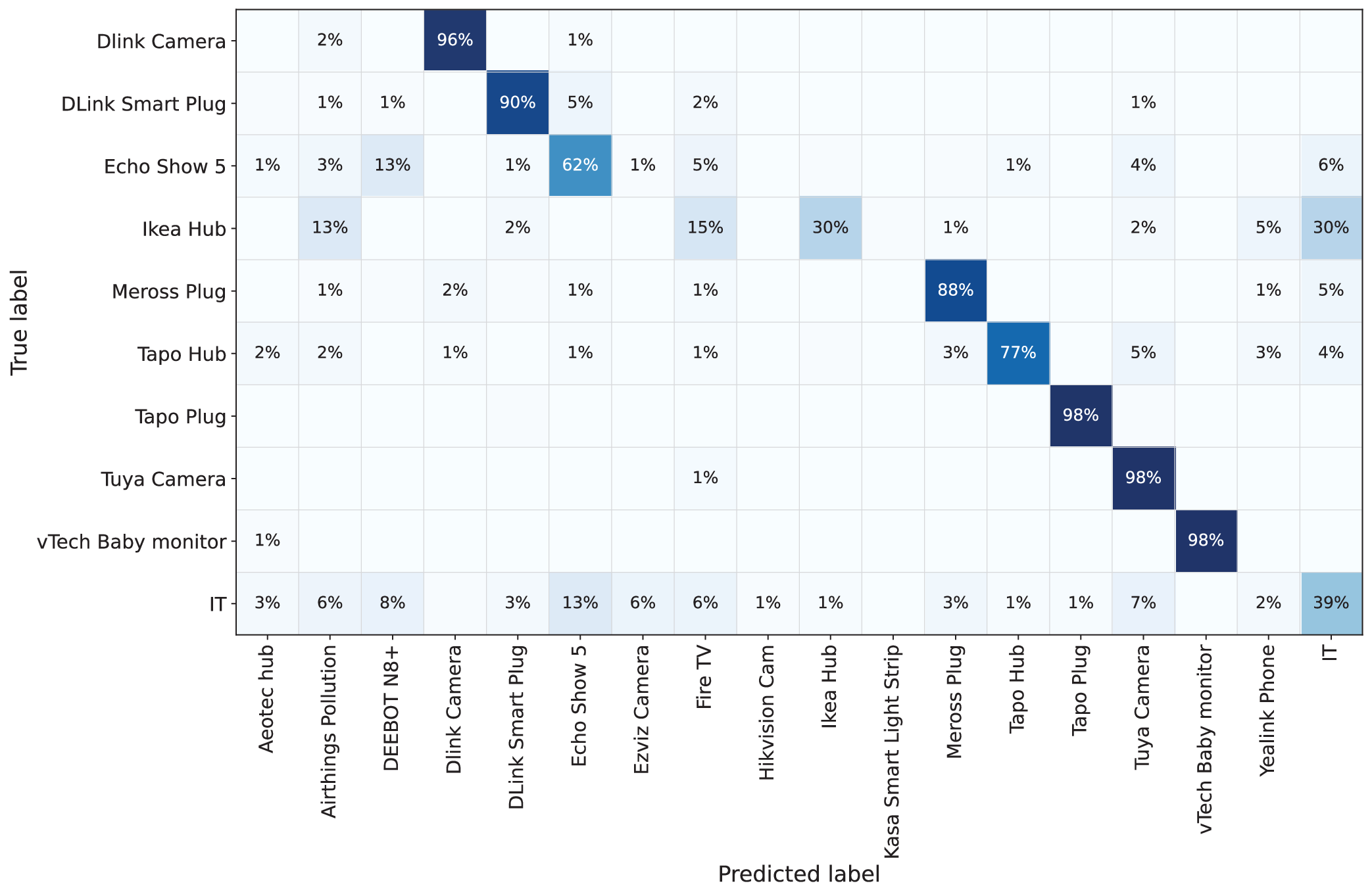}
    \caption{Confusion matrix of the downstream classifier trained on \Dataset{DATA25v1}{} embeddings generated by \EncoderVAE, evaluated on \Dataset{DATA25v2}{}.}
    \label{fig:mq-classification}
    \vspace{-6mm}
\end{figure}

To better understand how performance varies across device types under cross-environment deployment, Fig.~\ref{fig:mq-classification} shows the confusion matrix for the classifier using \EncoderVAE~embeddings, which achieves the highest macro F1-score among all evaluated encoders. We make two key observations from our cross-environment evaluation.

First, the performance gap between VAE-based and AE-based encoders widens substantially in the unseen environment. Both \EncoderVAE~and \EncoderVAELimited~outperform their AE counterparts by a large margin, indicating that latent regularization in VAEs improves robustness to distribution shift. 

Second, this experiment represents a stringent cross-environment test in which both the encoder and classifier are kept frozen, and no adaptation is performed. Despite differences in physical environment, user interactions and background traffic, six of the ten device types in \Dataset{DATA25v2}{} achieve recall exceeding $80$\%, indicating that their traffic representations transfer reliably across deployments. 

The remaining classes, namely Echo Show~5, Ikea Hub, Tapo Hub and \textit{IT}, exhibit lower recall ($62\%$, $30\%$, $77\%$ and $39$\%, respectively). These classes correspond to devices whose network behavior is inherently coupled to local context. In particular, the Ikea Hub and Tapo Hub function as application-layer gateways for Zigbee/Thread devices, such that their observed traffic reflects aggregated interactions with locally connected subdevices whose composition varies across environments. Similarly, Echo Show~5 frequently functions as a control hub for other devices, leading to environment-dependent communication patterns. The \textit{IT} class aggregates heterogeneous non-IoT devices operated by different users, resulting in high behavioral variability that limits cross-environment transfer without retraining.

These results show representation transfer without retraining; performance for these context-dependent classes can be further improved through retraining on the target environment.


\section{Conclusion}
Accurate identification of IoT devices in heterogeneous networks requires traffic analysis methods that can generalize beyond the conditions under which they are trained. 
In this paper, we showed that compact traffic representations learned from unlabeled IoT traffic using encoder--decoder models can achieve high reconstruction fidelity and remain stable across datasets. We further demonstrated that these high-quality representations can be used effectively for downstream device-type classification using simple classifiers trained on frozen encoders, indicating that most discriminative power lies in the learned embeddings rather than in classifier complexity.
Through a systematic benchmarking study cross-environment evaluation, we further showed that larger pretrained traffic encoders do not necessarily yield more robust representations for IoT device identification. In particular, representations learned with latent-space regularization exhibit substantially greater resilience to distribution shift than deterministic alternatives, even when evaluated under unseen deployment conditions. Using 18 million real IoT traffic flows collected across multiple years and environments, our findings highlight the importance of representation design and evaluation methodology for building practical and generalizable IoT traffic inference systems. To support reproducibility, we publicly release the datasets and encoder implementations used in this study.

\bibliographystyle{IEEEtran}
\bibliography{Gen-IoT-Class}
\balance

\end{document}